\documentclass[10pt,twocolumn]{article} 
\usepackage{simpleConference}
\usepackage{times}
\usepackage{amssymb}
\usepackage{times}  
\usepackage{helvet}  
\usepackage{courier}  
\usepackage[hyphens]{url}  
\usepackage{graphicx} 
\usepackage{natbib}  
\usepackage{caption} 
\usepackage{algorithm}
\usepackage{algorithmic}
\usepackage{stfloats} 
\usepackage[font=normalsize]{caption}

\usepackage{amsmath}  
\usepackage{amssymb}
\usepackage{multirow}  
\usepackage{pgfplots}
\usepackage{booktabs}
\usepackage{makecell}
\usepackage{subcaption}

\usepackage{newfloat}
\usepackage{listings}
\begin{document}

\title{Inter-Class Relational Loss for Small Object Detection: A Case Study on License Plates}

\author{Dian Ning, Dong Seog Han \\
\\
School of Electronic and Electrical Engineering \\
Kyungpook National University\\
Daegu, Republic of Korea\\
\\
ningdian@knu.ac.kr, dshan@knu.ac.kr \\
}

\maketitle
\thispagestyle{empty}

\begin{abstract}
In one-stage multi-object detection tasks, various intersection over union (IoU)-based solutions aim at smooth and stable convergence near the targets during training. However, IoU-based losses fail to correctly update the gradient of small objects due to an extremely flat gradient. During the update of multiple objects, the learning of small objects’ gradients suffers more because of insufficient gradient updates. Therefore, we propose an inter-class relational loss to efficiently update the gradient of small objects while not sacrificing the learning efficiency of other objects based on the simple fact that an object has a spatial relationship to another object (e.g., a car plate is attached to a car in a similar position). When the predicted car plate’s bounding box is not within its car, a loss punishment is added to guide the learning, which is inversely proportional to the overlapped area of the car’s and predicted car plate’s bounding box. By leveraging the spatial relationship at the inter-class level, the loss guides small object predictions using larger objects and enhances latent information in deeper feature maps. In this paper, we present twofold contributions using license plate detection as a case study: (1) a new small vehicle multi-license plate dataset (SVMLP), featuring diverse real-world scenarios with high-quality annotations; and (2) a novel inter-class relational loss function designed to promote effective detection performance. We highlight the proposed ICR loss penalty can be easily added to existing IoU-based losses and enhance the performance. These contributions improve the standard mean Average Precision (mAP) metric, achieving gains of 10.3\% and 1.6\% in mAP$^{\text{test}}_{50}$ for YOLOv12-T and UAV-DETR, respectively, without any additional hyperparameter tuning. Code and dataset will be available soon.
\end{abstract}

\section{Introduction}

Multi-object detection (MOD) is a widely used application that benefits from deep learning technologies, aiming to achieve both bounding box regression and accurate object classification. Depending on the size of the objects, they can be categorized as normal or small. Typical applications include vehicle license plate detection and aerial object detection. However, the small objects with limited pixel coverage~\cite{lin_microsoft_2015} have not been sufficiently addressed in the context of MOD. 

\begin{figure}[t]
\centering
\includegraphics[width=0.9\columnwidth]{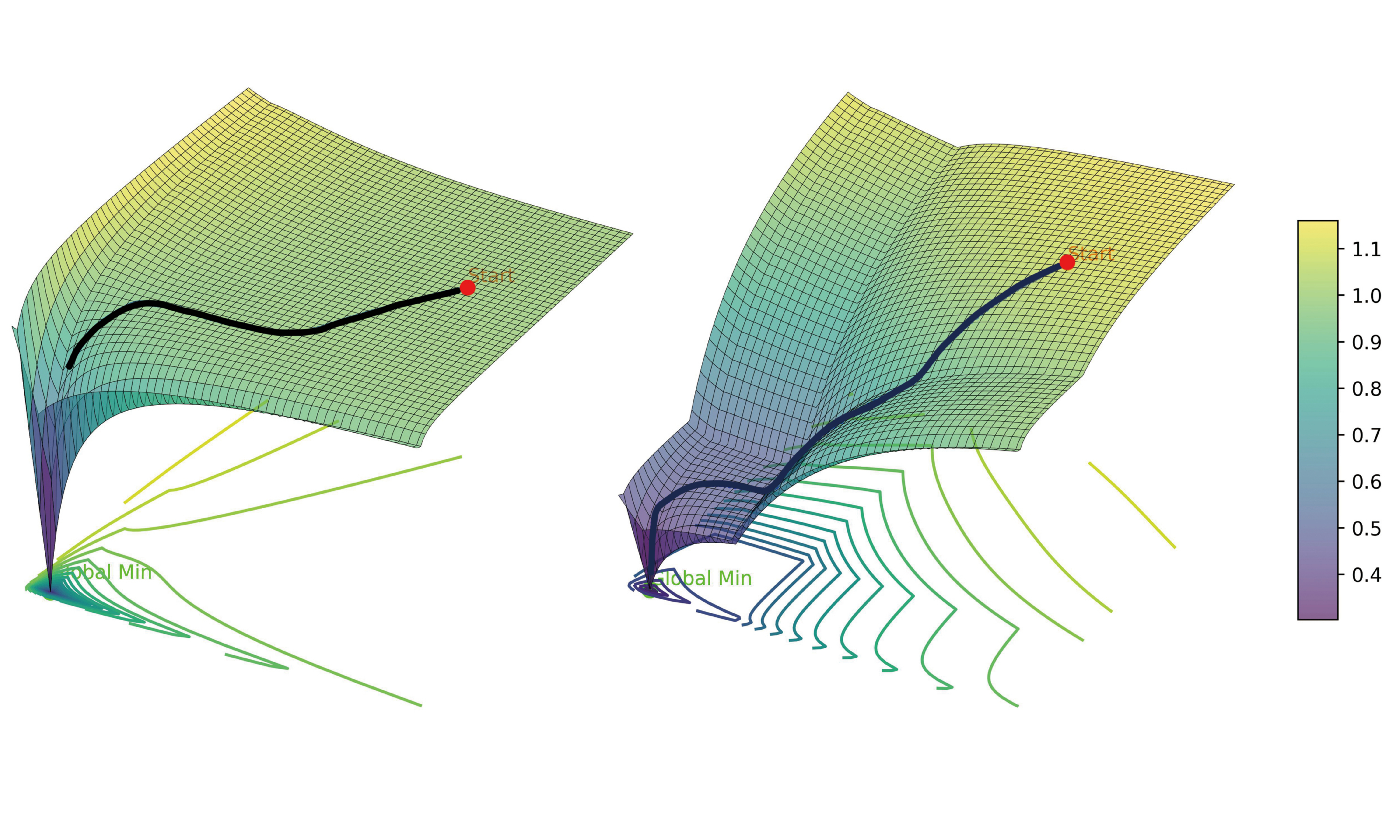}
\caption{The figure presents a comparison between the small object loss landscapes of CIoU loss (left) and CIoU loss plus inter-class relational (ICR) penalty (right) with the same settings. The projected contour lines reveal discrepancies in gradient density: the left plot displays a plateau with small gradient variance. In contrast, the right plot features denser contour lines converging toward the global minimum. The trajectories show gradient update over the same number of iterations. The right plot learns more efficiently by incorporating ICR relationships (e.g., license plate to car geometry), which guided the model towards a quicker and better regression path in the training phase. 
}
\label{fig1}
\end{figure}

Existing solutions generally extend the sophisticated designed object detection architectures, such as on you only looking once (YOLO) and the detection transformer (DETR). It can be categorized into two ways to handle small objects in the context of MOD tasks, small object detection (SOD) is used for the rest of the paper to represent this type of task. Model architecture modifications, such as diverse branch block~\cite{10474434} and high-resolution feature fusion~\cite{FFCA-YOLO}, can enhance the latent information in shallow layers and pass it to deeper layers, which introduces additional complexity to the model design and increases the parameter size. None architecture modification methods, such as loss design \cite{lin_focal_nodate}, parameter tuning \cite{DBLP:conf/cvpr/OuyangWZY16}, and data augmentation \cite{DBLP:journals/ict-express/SutramianiSS21}. These methods are generally effective for a specific dataset or model. Only a few insightful studies, such as focal loss~\cite{lin_focal_nodate}, commonly work for various models and datasets. 

Intersection over union (IoU)-based losses dominate the multi-object detections.
However, the IoU-based losses have critical bottlenecks in SOD, where the initial regression error is often small because of limited spatial features, which results in low gradient magnitudes and slow convergence (as shown in Fig.~\ref{fig1}, left). To correct the residual shape alignment, IoU-based losses may take larger gradient updates~\cite{10.5555/1953048.2021068, keskar2017largebatchtrainingdeeplearning} to overcome shape errors, which cause poor refinement and even loss oscillation~\cite{10205182}. Compared to the smooth gradient updates for normally sized objects, the excessively stringent gradient conditions~\cite{CFINet} for small objects often cause their predicted bounding boxes to be misclassified as negative samples~\cite{DBLP:journals/corr/abs-1908-05641}. This problematic gradient behavior makes the model disproportionately focus on larger objects that are easier to detect while neglecting small objects.

The loss design is efficient and lightweight if it is properly designed, like focal loss. In SOD, conventional IoU-based losses are inefficient, as mentioned above, leading to problems, such as insufficient sensitivity to the scale and location of small objects, inefficient gradient updates, especially when the predicted bounding box fits the ground truth (as shown in Fig.~\ref{fig:ciou_vs_icr}).
To resolve these issues, in SOD, such as car plate detection, based on an intuitive observation that a car plate is within an associated car form, we define this as a spatial relationship between objects. For example, a car (object $b$) always contains a car plate (object $a$) in a common case from a space perspective, and also a car plate (object $a$) is a component of the car (object $b$). Instead of focusing solely on achieving balance for individual classes, like focal loss, we explore the relationship between classes to guide small object learning in MOD. An extremely simple but effective inter-class relational loss penalty is defined to efficiently update the gradient of small objects while not sacrificing the learning efficiency of other objects. The ICR loss penalty is inversely proportional to the overlapped area of the car’s and the predicted car plate’s bounding box. By leveraging the spatial relationship at the inter-class level, the loss guides small object predictions by a larger object and also enhances the latent information of small object detection in deeper feature maps. The rest of the paper uses a vehicle license plate as a case study to illustrate the ICR loss penalty.

The contributions of this paper are summarized as follows:
\begin{itemize}
    \item Firstly, we design ICR loss using the inter-class spatial relationship in SOD. Our proposed loss gives an extra spatial rule, which provides fine-grained supervisory signals for small objects. Additionally, we highlight that the ICR loss penalty shows the potential across different models, datasets, and IoU-based losses. 
    \item Secondly, we create a new small vehicle multi-license plate dataset (SVMLP). The dataset consists of a one-to-one correspondence between a license plate and its corresponding vehicle. It includes 3,000 images with over 10,000 annotations in different backgrounds and light conditions. The annotations maintain high quality and generalization ability, which more adequately demonstrates the efficiency of the ICR loss penalty. 
\end{itemize}

\subsection{Small Object Detection} 

Recall that the SOD discussed in this paper is in the context of MOD. SOD aims to detect relatively small objects with tens to hundreds of pixels in an image \cite{xiuling_starting_2024}. In common objects in context (COCO), the size of the small objects is less than 32 $\times$ 32 pixels. With the advancement of modern imaging technologies, image resolution has significantly increased, leading to a reduction in the relative size of small objects within the image. Both small object detection dataset-A (SODA) \cite{cheng_towards_2023} and TinyPerson \cite{DBLP:conf/wacv/YuGJYH20} datasets define small objects based on their relative size in an image. For example, the average resolution of images in the SODA dataset is approximately 3,407 $\times$ 2,470 pixels; thus, objects smaller than 1,024 pixels are annotated as small. In this context, such small objects occupy less than 0.012\% of the total image area, posing significant challenges for accurate detection, which becomes more fatal in multi-object detection.

SOD algorithms can be divided into two mainstreams: two-stage and one-stage methods. The two-stage methods are initially proposed in region-based convolutional neural network (R-CNN) and faster R-CNN, which includes a region proposal network (RPN) to generate a region proposal and uses region of interest (RoI) to adjust classification. The architecture integrated visual geometry group (VGG)-based~\cite{cao_improved_2019} or residual neural network (ResNet)-based ~\cite{han_real-time_2019} deep backbones into RPN to extract features, resulting in degradation of small object details in deeper layers. 

To address the limitation of two-stage methods, one-stage methods combine classification and detection steps, defining object detection as a regression problem. The one-stage RetinaNet \cite{lin_focal_nodate} architecture uses a feature pyramid network (FPN) as backbone, which outperforms two-stage methods for the first time. Currently, one-stage architecture is mainly based on YOLO and DETR. YOLO uses CNNs as the backbone to extract features, a relatively lightweight model compared to DETR. SOD solutions usually improve the feature extraction in FPN and optimize the detection heads. On the other hand, DETR uses the transformer architecture to obtain contextual spatial information. However, CNN architecture, such as the ResNet baseline, is still used to encode the input image before entering the transformer, which loses some tiny region information~\cite{dubey_improving_2022,huang_dq-detr_2025}. Recent works try to use information augmentation methods after the CNN backbone, such as dynamic query selection~\cite{huang_dq-detr_2025} and feature fusion attention modules. 

In one-stage MOD, both YOLO and DETR use IoU-based loss. IoU~\cite{iouloss} is the common basic loss for evaluating the predicted bounding box in object detection. It uses the overlap area ratio of the prediction and target bounding boxes to calculate the loss. This ratio value is not sensitive to the scale or location between two boxes~\cite{cao_improved_2019}. The gradient will be zero when there is no intersection. 
The following popular IoU-based losses have been proposed to address these limitations with different measurements of IoU-based loss:
\begin{itemize}
    \item \textbf{Generalized IoU (GIoU)}~\cite{giouloss}, which handles the non-overlapping problem by incorporating the smallest enclosing convex box between two boxes, though it degenerates to IoU when boxes are nested;
    \item \textbf{Distance IoU (DIoU)}~\cite{DBLP:conf/aaai/ZhengWLLYR20}, which introduces the normalized distance between two box centers to encourage faster convergence;
    \item \textbf{Complete IoU (CIoU)}~\cite{ciouloss}, extends DIoU by considering the aspect ratio consistency between boxes, further improving localization accuracy for normal objects.
\end{itemize}
The distance centroids penalty of the predicted bounding box and the anchor is effective for normal objects. However, it is limited to the overall loss when dealing with small objects because the bounding boxes are small. Consequently, Soft-IoU \cite{ji_improved_2023} takes a penalty factor into account to reduce the influence of the ratio. Also, the normalized Wasserstein distance (NWD)~\cite{wang_normalized_2022} metric directly models the bounding box distribution as 2D Gaussian distributions, which improves the update of the gradient when there is no overlapping between bounding boxes. Focaler-IoU~\cite{zhang2024focalerioufocusedintersectionunion} dynamically amplifies the gradient contributions of small samples. 

\begin{table*}[!t]
\centering
\setlength{\tabcolsep}{5pt}  
\renewcommand{\arraystretch}{1.3} 
\caption{\textbf Attributes of SVMLP compared with UFPR-ALPR, CRPD and LSV-LP* \cite{10952375}.}
\begin{tabular}{l|c|c|c|c}
\hline
\textbf{ } & \textbf{UFPR-ALPR} & \textbf{CRPD} & \textbf{LSV-LP*} & \textbf{SVMLP} \\
\hline
Plates/Images &4,500/4,500  & 43,263/33,705 & 157k/145k  & 8,878/3,000 \\
\hline
No. Plates/Images &  1 & 1$\sim$7 (Avg: 1.28)  & 1$\sim$6 (Avg: 1.1)  & 1$\sim$10 (Avg: 2.96) \\
\hline
Min. absolute / relative plate area &   775px\textsuperscript{2}/ $3.4\times10^{-4}$  &  602px\textsuperscript{2}/ $3.1\times10^{-4}$ &  104px\textsuperscript{2}/ $5.3\times10^{-5}$ & 36px\textsuperscript{2}/ $6.0\times10^{-6}$ \\
\hline
Varying resolutions &  & \checkmark &  &  \checkmark  \\
\hline
Varying distances & Narrow & Narrow & Wide & Wide \\
\hline
Varying light conditions & day & day/night & day & day/night \\
\hline
Scenarios 
  & \makecell[l]{– Highway\\– Urban roads\\– Rural roads} 
  & \makecell[l]{– Cross road} 
  & \makecell[l]{– Highway\\– Urban roads\\- Parking} 
  & \makecell[l]{- Highway\\- Urban roads\\- Rural roads} \\
\hline
\end{tabular}
\label{tab:dataset_comparison_license}
\end{table*}

\subsection{Existing Vehicle and Vehicle License Plate}

Existing vehicle and vehicle license plate datasets can be divided into different practical scenarios: single-license plate datasets designed for parking systems and multiple-license plate datasets used in highway scenarios. However, most of these datasets contain only one bounding box per image or assume license plates are normally sized without fully considering the small object cases of license plates in real-world conditions. In practice, license plates are often captured at long distances or under non-frontal angles, which makes detection more challenging. In this section, we introduce several representative public datasets and highlight their key characteristics in the context of SOD.

\textbf{UFPR-ALPR dataset}, a multi-label dataset~\cite{ufpr—alpr} contains 4,500 images with 30,000 LP labels in varIoUs urban environments. The images were captured in Brazil with a frame size of 1,920 $\times$ 1,080 pixels. UFPR-ALPR includes different types of vehicles, such as sedans, motorcycles, and public vehicles with different sizes of Brazilian LPs. In this dataset, the researchers did not label all the LPs and divided them into visible and invisible, which only labelled the visible LP in the foreground.

\textbf{CRPD dataset}, a Chinese road plate dataset~\cite{crpd} contains over 30,000 annotated images, which consist of images captured by surveillance perspective at a fixed angle. This dataset is suitable for traffic monitoring systems as the images were captured using electronic monitoring systems. The dataset consists of about 80\% images with a single license plate label and 20\% images with two to three license plate labels.

\textbf{LSV-LP dataset}, a large-scale video-based license plate dataset~\cite{lsplp} contains 401,347 video frames and 364,607 annotated license plates. It labels multiple license plates and vehicles in one image. The innovative combination of data from dynamic and static shots resulted in three sub-datasets: move vs. move, move vs. static, and static vs. move, which depend on the motion states of the camera and the vehicles. This subsets provide different real-world scenarios for evaluations. The LSV-LP dataset is large-scale and diverse, however, it contains a non-negligible amount of noisy data \cite{DBLP:journals/tits/PengGMX24}.

\section{The SVMLP Dataset}
To solve the noisy data and lack of different size of license plate problem, we propose a dataset, named the small vehicle multi-license plate dataset (SVMLP), to find the relationship between a car and its license plate in real-world scenarios. In this section, we introduce SVMLP from its source, collection methods, and annotation methods, and comparison with other datasets. 

\subsection{Collection and Annotation} 
\textbf{Source} We obtained high-resolution in-car footage videos from YouTube with the authors' explicit authorization under a non-commercial license. Video scenes cover several important features for license plate detection tasks in terms of different cities, road types, and brightness. Chinese cities were selected to make sure the basic design of car plates is similar, such as Shenzhen, Meizhou, Shenyang, and Chongqing, etc. Road type includes downtown and highways, which were recorded during daytime and nighttime with the field of view. 

\textbf{Collection} We collected approximately 10 hours of video from the authorized YouTube source. To prepare the dataset, we extracted frames from the videos at a sampling rate of 60 frames per second. Images were available in JPG format and ranged in size from 1,080 $\times$ 1,280 to 1,920 $\times$ 1,080 pixels. Some frames in the extracted image frames were not useful because of duplication or high similarity of annotations. Thus, we selected different images with a random sample frequency from the time domain to ensure that every selected image contains different background information, which makes the dataset enriched and robust.

\textbf{Labeling} SVMLP emphasizes the inter-class level relationship between the vehicle and its license. Each vehicle and its corresponding license plate are treated as a pair, ensuring that each license plate is within its associated vehicle bounding box. The label was managed as YOLO-style annotation files. We organized each set such that the first line indicates the license plate with class index 0, and the second line corresponds to the vehicle associated with class index 1. This annotation structure is easy to be used for training and evaluation tasks in YOLO series.

\subsection{Dataset Analysis}
SVMLP includes 3,000 images divided into train, validation, and test sets, which contain 2,400, 300, and 300 images, respectively. Each subset contains an equal number of images in daytime and nighttime scenarios. Moreover, the vehicle and its corresponding license plate are labeled in each image, which means there is no individual license plate or vehicle missing its paired class label. We annotated 8,878 license plates and 8,878 vehicles, totaling 17,756 objects. The annotations of vehicles may overlap with their surrounding vehicles. 

To study the distribution of our dataset, the definition of object size in COCO dataset is used to classify small, medium, and large objects. According to the definition, 4,670 license plates are classified as small, and 3,852 as medium. The vehicle bounding boxes are 1093 of medium and 7785 of large sizes. 

\subsection{Compare with existing datasets}
Our proposed SVMLP dataset offers several key advantages over existing datasets, such as UFPR-ALPR, CRPD, and LSV-LP*. The summary is shown in Table.~\ref{tab:dataset_comparison_license}. For LSV-LP*, only the move to move subset is used for analysis, and thus it is marked with an asterisk (*). First, SVMLP includes an average of 2.96 license plates per image, encouraging multi-object detection. Second, it includes the smallest annotated license plates in both absolute and relative size of small objects. Furthermore, SVMLP dataset is highly diverse in terms of image resolution, object scale, lighting condition (day and night), and scenarios (highways, urban roads, and rural roads). This comprehensive variation is ideal for robust training and evaluation under real-world conditions.

\begin{figure}[!h]
\centering
\includegraphics[width=\columnwidth]{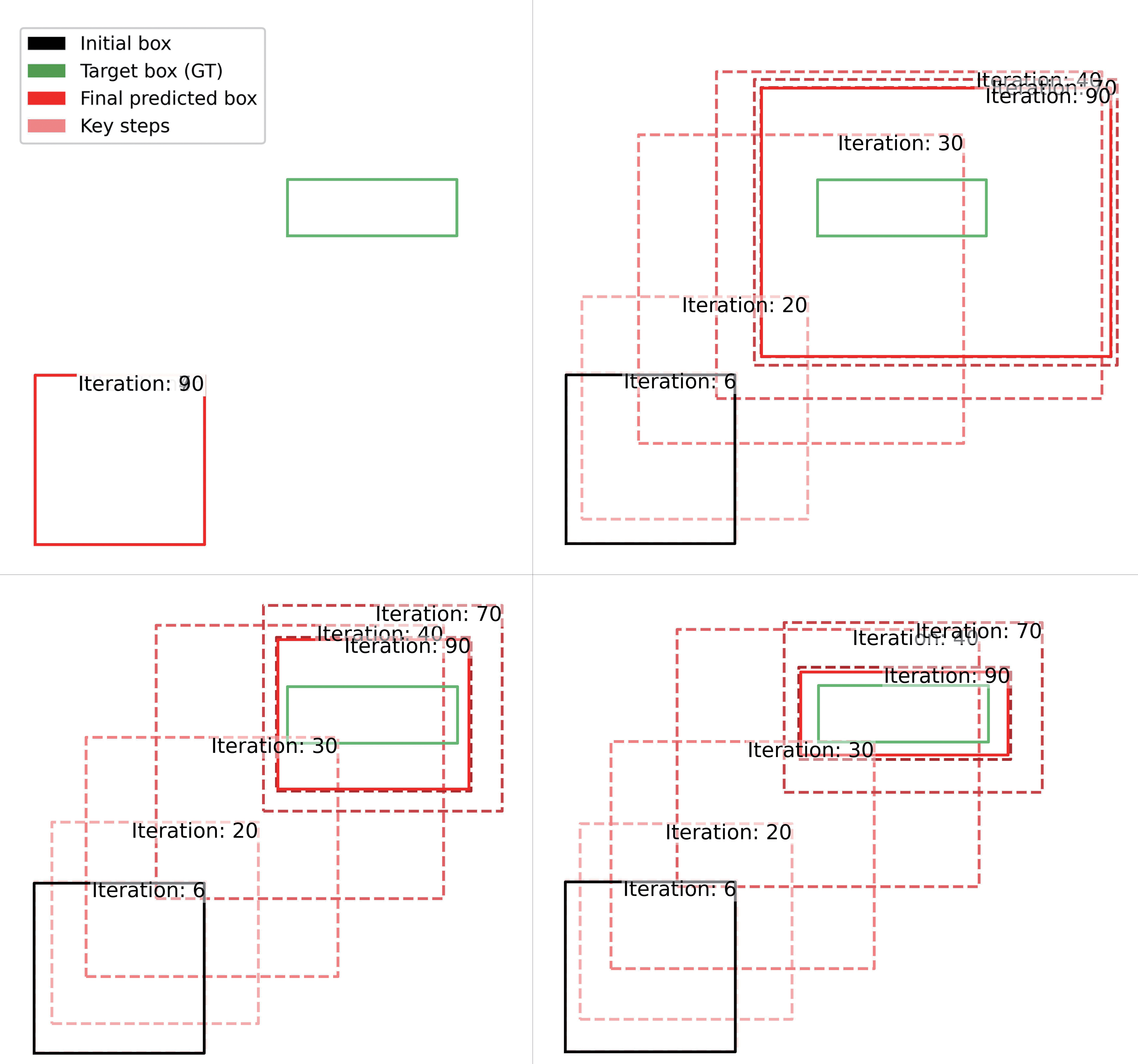}
\caption{
Visualization of IoU (top-left), GIoU (top-right), DIoU (bottom-left), and CIoU (bottom-right) losses in 100 iterations. }
\label{fig:problem state}
\end{figure}

\section{Methodology}

\subsection{Problem Statement}
IoU-based losses aim to establish a one-to-one correspondence between the ground truth and predicted object, however, they are insufficient to fully capture the discrepancies. IoU-based losses generally considering geometric factors: overlapping area, distance, and aspect ratio, which are defined as $\mathcal{S}(\mathcal{B}, \mathcal{B}_{\text{gt}})$, $\mathcal{D}(\mathcal{B}, \mathcal{B}_{\text{gt}})$, and $\mathcal{V}(\mathcal{B}, \mathcal{B}_{\text{gt}})$, respectively. The $\mathcal{B}$ is predicted bounding box and $\mathcal{B}_{\text{gt}}$ is its target bounding box. They can be formulated by different metrics, we only discuss the limitations of geometric factors, such as overlapping area, distance, and aspect ratio, which are denoted by $\mathcal{S}$, $\mathcal{D}$, and $\mathcal{V}$, respectively.

Fig.~\ref{fig:problem state} shows the different IoU-based loss discrepancies between the predicted small license plate bounding box and its corresponding ground truth. 
IoU loss cannot handle the case when there is no overlap between the predicted bounding box and the ground truth, which only uses $\mathcal{S}$. GIoU also only considers $\mathcal{S}$, the non-overlapping can be alleviated using a different calculation. However, $\mathcal{S}$ lacks shape constraints, the predicted box tends to predict larger boxes after multiple iterations, which may lead to convergence at a local minimum. DIoU adds the distance factor $\mathcal{D}$ where the gradient of the predicted bounding box can move to the targets. However, its contribution to gradient updates becomes limited due to the normalization factor in the denominator. CIoU considers both $\mathcal{S}$, $\mathcal{D}$, and $\mathcal{V}$. However, when the target is small, the gradient update is insufficient, thus, it needs more iterations. It cost 70 iterations to move to the target's surroundings as shown in Fig.~\ref{fig:problem state}.

It is very common in model training because initial boxes have similar values. Predictions are often randomly shaped in early training, which may not significantly contribute to training. Therefore, these far-distant predicted bounding boxes have very similar scores, which hinders effective gradient updates. Therefore, how to quickly and correctly guide the initial bounding boxes to the targets is essential in SOD.

\subsection{Solution Design}
To address the problems of gradient update in SOD, We propose an IoU penalty called the inter-class relation penalty. It uses contextual information from correlated large objects to refine the step size of the gradient. A simplest form of ICR loss penalty is that if the predicted bounding box of license plate is not within the ground truth bounding box of vehicles, a coefficient $\delta$ is multiplied to any IoU-based losses as defined in equation~\ref{eq:eq1}, where $\delta > 1$ and is suggested to be different values depending on different datasets and models.
\begin{equation}
\label{eq:eq1}
    \mathcal{L}_{\text{ICR-CIoU}} = \delta\mathcal{L}_{\text{IoU-based}}, \delta > 1
\end{equation}

\begin{figure}[!t]
\centering
\includegraphics[width=\columnwidth]{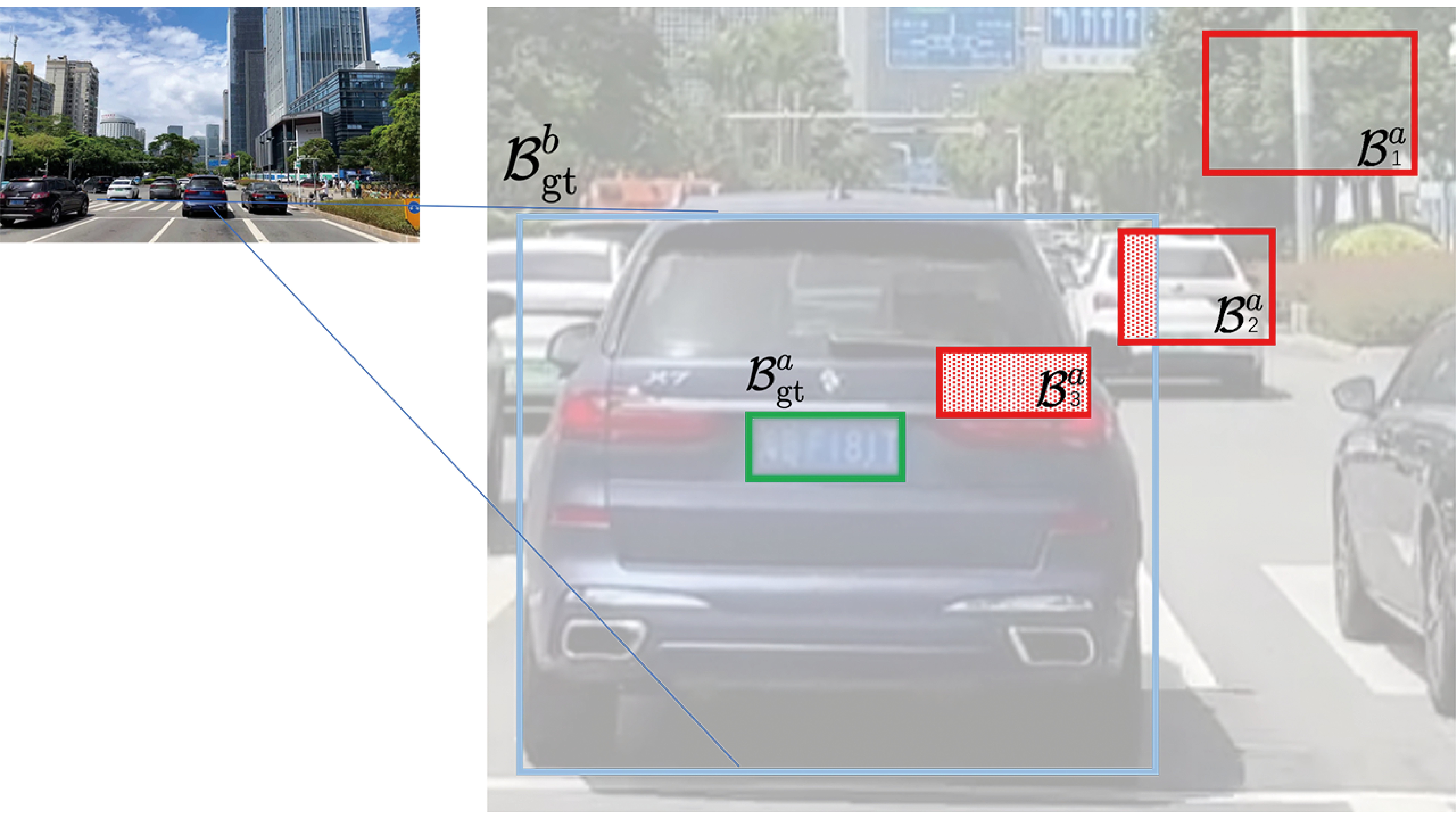}
\caption{
Illustration of the relationship between three $\mathcal{B}^{a}$ (red), $\mathcal{B}^{a}_\text{gt}$ (green), and $\mathbf{B}_{\text{gt}}^b$ (blue).}
\label{fig:solution}
\end{figure}

To better address the efficient gradient update of SOD, we borrow the methodology of IoU and combine it with the proposed inter-class relation. The larger object is defined as the area that has a relatively fixed spatial relationship to the small targets, as shown in Fig.~\ref{fig:solution}.
The set of predicted bounding boxes for the targeted small objects $a$ is denoted by $\mathbf{B}^a = [\mathbf{B}_1^a, \mathbf{B}_2^a, \dots, \mathbf{B}_N^a]^{\top} \in \mathbb{R}^{N \times 4}$, where $\top$ denotes the transpose operation. The ground truth of license plate and vehicle are denoted by $\mathbf{B}_{\text{gt}}^a$ and $\mathbf{B}_{\text{gt}}^b \in \mathbb{R}^{N \times 4}$, respectively. The overlapping area between $\mathcal{B}^{a}$ and its corresponding $\mathcal{B}^{b}{\text{gt}}$ is normalized by the area of the predicted bounding box $\mathcal{B}^{a}$ is used to indicate inter-class relation, which is calculated by

\begin{equation}
\mathcal{R}_{\text{ab}} =\frac{\mathcal{B}^{a} \cap \mathcal{B}^{b}_{\text{gt}}}{\mathcal{B}^{a}}.
\end{equation}

Based on the spatial relationship between two classes, the new factor is 
\begin{equation}
\label{eq:eq33}
\mathcal{L}_{\text{ICR-IoU}} = \left(\delta \left(1 -\mathcal{R}_{\text{ab}} \right) +1\right)\mathcal{L}_{\text{IoU-based}},
\end{equation}
where $\mathcal{L}_{\text{IoU-based}}$ can be replaced by other IoU-based loss. For example, $\mathcal{L}_{\text{ICR-CIoU}}$ is calculated using euqation~(\ref{eq:eq33}) by replacing $\mathcal{L}_{\text{IoU-based}}$ with CIoU loss. $\delta$ is the weight of this penalty, which is 2.5 by default as illustrated in Fig.~\ref{fig:impact_factor}. $\mathcal{L}_{\text{ICR-IoU}}$ is calculated based on $\mathcal{B}^{b}_{\text{gt}}$, therefore, the penalty increases when the predicted small object deviates from its associated large object region. 
Meanwhile, in this paper, the proposed method are applied to popular IoU-based loss functions, which are $\mathcal{L}_{\text{ICR-IoU}}$, $\mathcal{L}_{\text{ICR-DIoU}}$, and $\mathcal{L}_{\text{ICR-GIoU}}$.

\section{Experiments}
In this section, extensive simulations are conducted to perform quantitative analysis, which demonstrates the robustness and effectiveness of the proposed inter-class loss compared to IoU-based methods. 

\subsection{Experimental Settings}
Four datasets, UFPR-ALPR, CRPD, LSV-LP, and proposed SVMLP, are used for evaluation in various scenarios. These public datasets are partitioned according to their officially recommended train, validation, and test splits.

Two mainstream baselines designed for one-stage detectors are used to evaluate ICR loss: YOLO series and DETR series. The YOLOv9-t detector with CIoU loss serves as a baseline and is trained on four datasets. To verify the effectiveness of the ICR loss with commonly used IoU-based losses, we also conduct experiments using YOLOv9-t with different loss functions, such as IoU, DIoU, and GIoU losses. 

In YOLO series, the models were trained with an input resolution of 640 $\times$ 640, a batch size of 32, and an initial learning rate of 0.01 using a stochastic gradient descent (SGD) optimizer. Training was conducted for 100 epochs on 3 NVIDIA RTX 4090 GPUs. The default anchor settings of YOLOv9 are adopted. In DETR series, we use the same input resolution, hardware, and software environment. The majority of settings are the same as the original settings in YOLO and DERT. A few data augmentation functions are disabled due to training efficiency, details are in the appendix.

\begin{table*}[!t]
\centering
\caption{\textbf{Ablation study on different baseline models} w/o $\mathcal{L}_{\text{ICR}}$ based on SVMLP validation and test sets.}
\setlength{\tabcolsep}{2pt}
\begin{tabular}{l c c c c | c c c c | c c c c}
\textbf{Model} & \textbf{Backbone} & \textbf{\#Epochs} & \textbf{\#Params (M)} & \textbf{$\mathcal{L}_{\text{ICR}}$} & \textbf{AP$^{\text{val}}_{50}$} & \textbf{AP$^{\text{val}}_{50:95}$} & AP$^{\text{val}}_{S}$ & AP$^{\text{val}}_{M}$   & \textbf{AP$^{\text{test}}_{50}$} & \textbf{AP$^{\text{test}}_{50:95}$} & AP$^{\text{test}}_{S}$ & AP$^{\text{test}}_{M}$ \\
\toprule       
\multicolumn{13}{l}{\textit{YOLO Detectors}} \\
\toprule       
\multirow{2}{*}{YOLOv9-T \cite{DBLP:conf/eccv/WangYL24}} 
    & \multirow{2}{*}{-} & \multirow{2}{*}{100} & \multirow{2}{*}{2.6} &            & 60.6 & 37.3 & 12.3 & 55.7  & 60.2 & 37.5 & 12.1 & 62.3  \\
    &                    &                      &                       & \checkmark & \textbf{66.7} & \textbf{41.2} & \textbf{16.5} & \textbf{57.4} & \textbf{65.6}  & \textbf{40.4} &  \textbf{15.5} & \textbf{63.6}\\
\midrule       
\multirow{2}{*}{YOLOv9-S} 
    & \multirow{2}{*}{-} & \multirow{2}{*}{100} & \multirow{2}{*}{9.6} &             & 72.1  & 43.9  & 19.0  & 56.9  & 71.3  & 43.6 &  19.0 &64.2\\
    &                    &                      &                       & \checkmark & \textbf{75.1} & \textbf{45.7} & \textbf{22.4} & \textbf{58.8} & \textbf{74.7} & \textbf{46.2}  & \textbf{21.8} &\textbf{65.9} \\
\midrule       
\multirow{2}{*}{YOLOv9-M} 
    & \multirow{2}{*}{-} & \multirow{2}{*}{100} & \multirow{2}{*}{32.5} &             & 70.6  & 46.9  & 19.1  & 60.1  & 70.6  & 46.5 & 18.8 &69.2 \\
    &                    &                      &                       & \checkmark & \textbf{74.1} & \textbf{49.4} & \textbf{22.7} & \textbf{60.9} & \textbf{74.1} & 49.4 &\textbf{22.5} & 69.0 \\
\midrule       
\multirow{2}{*}{YOLOv9} 
    & \multirow{2}{*}{-} & \multirow{2}{*}{100} & \multirow{2}{*}{60.5} &             &  73.2  & 48.0 & 21.4  & 60.9  & 73.6 & 48.7 & 22.0 & 69.1 \\
    &                    &                      &                       & \checkmark & \textbf{74.6} & \textbf{49.3} & \textbf{22.2} & \textbf{62.1} & \textbf{74.6} & \textbf{50.0} & 22.0 &  \textbf{70.3}\\
\midrule       
\multirow{2}{*}{YOLOv11-T \cite{DBLP:journals/corr/abs-2410-17725}} 
    & \multirow{2}{*}{-} & \multirow{2}{*}{100} & \multirow{2}{*}{2.5} &             & 61.3  &37.4  & 10.6  & 50.5  & 61.3  & 37.9 & 10.4 & 57.1 \\
    &                    &                      &                       & \checkmark & \textbf{72.2} & \textbf{44.8} & \textbf{18.1} & \textbf{53.6} & \textbf{72.6} & \textbf{45.2} & \textbf{17.2}& \textbf{61.7} \\
\midrule       
\multirow{2}{*}{YOLOv12-T \cite{DBLP:journals/corr/abs-2502-12524}} 
    & \multirow{2}{*}{-} & \multirow{2}{*}{100} & \multirow{2}{*}{2.6} &             & 62.2  & 37.3  & 11.1  &  50.8 & 61.1  & 37.4  & 10.3& 57.4\\
    &             &           &                       & \checkmark & \textbf{71.1} & \textbf{43.7} & \textbf{16.9} & \textbf{62.0} & \textbf{71.4} & \textbf{44.0} & \textbf{16.9} & \textbf{62.0}\\
\hline
\multicolumn{13}{l}{\textit{DETR Detectors}} \\
\hline
\multirow{2}{*}{RE-DETRv2 \cite{DBLP:journals/corr/abs-2407-17140}} 
    & \multirow{2}{*}{R18} & \multirow{2}{*}{72} & \multirow{2}{*}{20.0} &             & 92.8 & 58.8    & 47.3    & 63.7  & 93.1   & 61.1 & 51.2& 71.3    \\
    &                      &                      &                       & \checkmark & \textbf{93.5} & \textbf{59.6}  & \textbf{47.5}    & \textbf{64.1}  & \textbf{93.6} & \textbf{61.4} & 50.7 & 70.9 \\
\midrule       
\multirow{2}{*}{RE-DETRv2} 
    & \multirow{2}{*}{R34} & \multirow{2}{*}{72} & \multirow{2}{*}{31.0} &             & 91.6 & 56.6    & 44.7    &61.8  & 91.7   & 58.2 & 46.8& 69.6    \\
    &                      &                      &                       & \checkmark & 91.0 & \textbf{56.9}  & \textbf{45.1}    & \textbf{63.1}  & \textbf{92.8} & \textbf{60.1} & \textbf{49.2} & 68.5 \\
\midrule       
\multirow{2}{*}{UAV-DETR \cite{DBLP:journals/corr/abs-2501-01855}} 
    & \multirow{2}{*}{R18} & \multirow{2}{*}{100} & \multirow{2}{*}{21.2} &             & 91.1  & 56.2    &43.3   &  63.1  & 93.1   & 59.1   & 45.9 &71.8\\
    &                      &                      &                        & \checkmark & \textbf{92.3} & \textbf{57.1}   & \textbf{44.9}    & \textbf{63.4} & \textbf{94.7}    & \textbf{60.4} & \textbf{48.4} & \textbf{72.3}  \\
\midrule       
\multirow{2}{*}{UAV-DETR } 
    & \multirow{2}{*}{R50} & \multirow{2}{*}{100} & \multirow{2}{*}{44.6} &             & 93.9 &  58.4   & 45.3  & 64.6  &95.0   & 61.8   & 50.0 &74.5  \\
    &                      &                      &                        & \checkmark  & \textbf{94.4}  &  \textbf{59.0}   & \textbf{46.5}   & 64.6  & \textbf{96.0}   & \textbf{62.5}   & \textbf{51.2} & 74.2  \\
\hline

\end{tabular}
\label{tab:val_diff_model}
\end{table*}

\begin{table*}[!t]
\centering
\caption{\textbf{Comparison results of $\mathcal{L}_{\text{ICR}}$ for objects A (license plate) and B (vehicle) across datasets}, along with the AP metrics of objects A, B, small and medium-sized objects. Metrics are reported on YOLOv9-T validation and test evaluation.}
\setlength{\tabcolsep}{6pt}
\begin{tabular}{l | c | cc cc c c | cc cc c c}
\hline
\textbf{Dataset} & $\mathcal{L}_{\text{ICR}}$ 
& \multicolumn{2}{c}{AP$^{\text{val}}_{50}$} 
& \multicolumn{2}{c}{AP$^{\text{val}}_{50:95}$} 
& AP$^{\text{val}}_{S}$ & AP$^{\text{val}}_{M}$ 
& \multicolumn{2}{c}{AP$^{\text{test}}_{50}$} 
& \multicolumn{2}{c}{AP$^{\text{test}}_{50:95}$} 
& AP$^{\text{test}}_{S}$ & AP$^{\text{test}}_{M}$ \\
& & A & B & A & B &  &  & A & B & A & B &  &  \\
\hline
\multirow{2}{*}{SVMLP} 
    &                  & 60.6 & 97.5 & 37.3 & 81.4 & 12.3 & 55.7 & 60.2 & 97.0 & 37.5 & 88.7 & 12.1 & 62.3\\
    & \checkmark       & \textbf{66.7} & 97.4 & \textbf{41.2} & 
        \textbf{81.5} &\textbf{16.5} & \textbf{57.4} & \textbf{65.6} & \textbf{97.2} & \textbf{40.4} & \textbf{88.6} & \textbf{15.5} & \textbf{63.6} \\

\hline
\multirow{2}{*}{UFPR-ALPR} 
    &                  & 98.5 & 99.4 & 64.0 & 90.3 & 56.8 & 59.8 &97.3&  99.5  &  63.9   &  90.2  &  56.8  &61.4     \\
    & \checkmark       &  \textbf{99.5} &  \textbf{99.5} &  \textbf{68.7} & \textbf{93.1} & \textbf{65.9} & \textbf{64.6} &  \textbf{98.8}  &  95.5   &  \textbf{69.4}   &   \textbf{91.4}  &  \textbf{61.4}   &    \textbf{67.2}  \\
\hline
\multirow{2}{*}{CRPD} 
    &                  & 99.0 & 98.9 & 81.9 & 94.2 & 4.3 & 79.4 & 98.4  & 98.4 &  83.8&  93.4 &  5.3   & 81.7 \\
    & \checkmark       & 99.0 & 98.9 & \textbf{82.3} & 94.2& \textbf{13.1} & \textbf{79.8} & 98.4    &  98.3   &  \textbf{84.1}   &   93.3  &  5.0   &   \textbf{82.0}  \\
\hline
\multirow{2}{*}{LSP-LP*} 
    &                  & 82.8 & 87.4 & 51.1 & 53.6 &25.1 & 35.0 & 86.7 & 90.4 & 51.6 & 56.0 &      10.6 & 31.1   \\
    & \checkmark       & \textbf{83.1} & 85.1 & \textbf{52.9} & 52.3 & \textbf{26.5} & 32.0 & \textbf{88.5} & 88.6 & \textbf{55.6} & 54.8 & \textbf{12.3}    &  \textbf{34.0}   \\
\hline
\end{tabular}
\label{tab:dataset_withdiffdataset}
\end{table*}

\subsection{Experimental Results} 

Table~\ref{tab:val_diff_model} shows the performance of the proposed $\mathcal{L}_{\text{ICR}}$ compared to various baseline models, including YOLO-based and DETR-based architectures based on the SVMLP dataset. Consistent improvements are observed across all models, especially in average precision (AP) for small objects $\text{AP}^{\text{val}}_{S}$.

For YOLO models, applying $\mathcal{L}_{\text{ICR-CIoU}}$ enhances both the overall AP and the small-object AP on the validation set. On the test set, our method demonstrated compatibility with recent YOLO versions (v9 to v12) and achieved a better performance, especially on YOLOv12-T. Specifically, it achieves a 10.3\% increase in AP50 and 6.6\% in AP50:95, while increasing the small object AP from 10.3\% to 16.9\%. Similar trends can be found in large models, such as YOLOv11-T, where the small object AP is improved by up to 6.8\%. Despite the faster early-stage convergence of larger models due to their higher capacity, our loss function consistently improves all YOLO versions during gradient updates.
In DETR models, although the original method leveraged the Hungarian matching algorithm to achieve high AP, our approach still yielded 0.5\%$\sim$2.5\% absolute improvements. For instance, UAV-DETR with a ResNet-50 backbone improves the small object AP from 50.0\% to 51.2\%, and achieves a 1.0\% absolute gain in AP50 on the test set.

In Table~\ref{tab:dataset_withdiffdataset} presents the ablation results of \(\mathcal{L}_{\text{ICR-CIoU}}\) across four datasets. Based on different datasets, \(\mathcal{L}_{\text{ICR}}\) outperforms the CIoU loss in terms of most AP and average recall (AR) metrics on the same YOLOV9-T model. For SVMLP dataset, $\mathcal{L}_{\text{ICR-CIoU}}$ increases the AP50 score from 60.6\% to 66.7\% and the AP50:95 score from 37.3\% to 41.2\% for object A (license plate) in the validation dataset. Similar improvements are observed on the test set, where AP50 increases from 60.2\% to 65.6\%, and AP50:95 increases by 2.9\%, validating its effectiveness in improving generalization. The AP for small and medium objects increases by 4.3\% and 3.4\%, respectively, with the proposed method in both sets.

There are three important conclusions based on the above tables: 1) The proposed inter-class loss effectively improves both AP and AR on validation and test sets, especially for small objects. On the UFPR-ALPR test set, our method achieves a 1.2\% to 1.5\% increase in AP, increases small-object AP by 4.6\%, and increases medium-object AP by 9.5\%; 2) The performance of object B (vehicle) remains stable under the multi-class detection; 3) The performance is most pronounced when training data is accurately labeled. If the used dataset has significant label noise or dirty annotations, the improvement becomes less stable, as the performance shown in CRPD and LSV-LP.

Table~\ref{tab:val_ICR_ab} shows the quantitative comparison of models trained with $\mathcal{L}_{\text{ICR-IoU}}$, $\mathcal{L}_{\text{ICR-DIoU}}$ and $\mathcal{L}_{\text{ICR-GIoU}}$. The proposed ICR loss penalty can reliably improve small objects' AP metrics based on existing IoU-based losses. For instance, $\mathcal{L}_{\text{ICR-IoU}}$ significantly improves AP metrics without increasing any model parameters. It is the fact that CIoU provides more informative gradient feedback by jointly considering overlap area, distance, and aspect ratio, which helps better localize small objects. In this case, our inter-class loss further enhances the steps between predictions and targets by penalizing directionally inconsistent updates. Thus, it becomes more effective when the base loss has a larger step size and updates the gradient softly. In contrast, for simpler losses like IoU or GIoU, the relatively rough gradient supervision limits the expressiveness of $\mathcal{L}_{\text{ICR}}$, resulting in moderate gains.

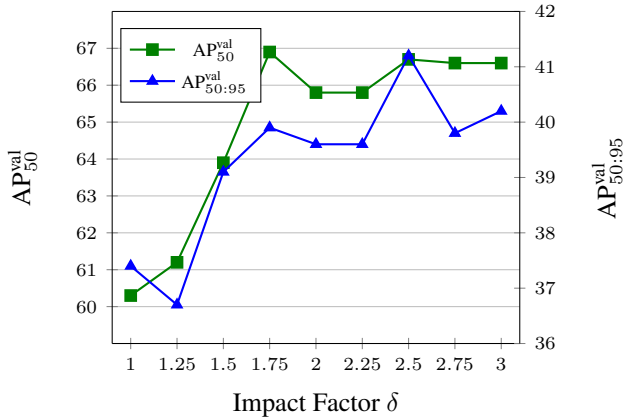
\begin{figure}[ht]
\centering
\begin{tikzpicture}
\begin{axis}[
    width=7cm, height=6cm,
    axis y line*=left,
    axis x line=bottom,
    xlabel={Impact Factor $\delta$},
    ylabel={AP$^{\text{val}}_{50}$},
    ylabel style={yshift=-1pt},
    ymin=59, ymax=68,
    xmin=0.9, xmax=3.1,
    xtick={1.0,1.25,1.5,1.75,2.0,2.25,2.5,2.75,3.0},
    ytick={60,61,62,63,64,65,66,67},
    axis y line=left,
    axis line style={-}, 
    ymajorgrids=true,
    enlargelimits=false,
    legend style={
        at={(0.02,0.95)},
        anchor=north west,
        font=\scriptsize,
        fill=white,
        draw=black
    },
    tick label style={font=\scriptsize},
]

\addplot[
    color=green!50!black,
    mark=square*,
    thick,
] coordinates {
    (1.0,60.3) (1.25,61.2) (1.5,63.9) (1.75,66.9)
    (2.0,65.8) (2.25,65.8) (2.5,66.7)
    (2.75,66.6) (3.0,66.6)
};

\addlegendentry{$\text{AP}^{\text{val}}_{50}$}
\addplot[
    color=blue,
    mark=triangle*,
    thick,
] coordinates { (1.0,37.4)};
\addlegendentry{$\text{AP}^{\text{val}}_{50{:}95}$}

\end{axis}

\begin{axis}[
    width=7cm, height=6cm,
    axis y line*=right,
    axis x line=bottom,
    xlabel={},
    ylabel={$ \text{AP}^{\text{val}}_{50{:}95} $},
    ylabel style={
        yshift=0pt,
        xshift= -2pt,
    },
    ymin=36, ymax=42,
    xmin=0.9, xmax=3.1,
    ytick={36,37,38,39,40,41,42},
    axis y line=right,
    axis x line=none,
    axis line style={-},  
    tick label style={font=\scriptsize},
    ymajorgrids=false,
]

\addplot[
    color=blue,
    mark=triangle*,
    thick,
] coordinates {
    (1.0,37.4) (1.25,36.7) (1.5,39.1)
    (1.75,39.9) (2.0,39.6) (2.25,39.6)
    (2.5,41.2) (2.75,39.8) (3.0,40.2)
};

\end{axis}

\draw[black, line width=0.3pt] (rel axis cs:0,1) -- (rel axis cs:1,1);

\end{tikzpicture}
\caption{\textbf{Performance under different impact factor $\delta$} in $\mathcal{L}_{\text{ICR}}$ on the SVMLP validation set using the YOLOV9-T.}
\label{fig:impact_factor}
\end{figure}

Different $\delta$ can affect the gradient update, as shown in Fig.~\ref{fig:impact_factor}. The AP remains relatively stable within the range of 1.75 to 3.0, with the best performance achieved at $\delta=2.5$, resulting in approximately a 1\% improvement in both AP50:95 and AP50.

\begin{table}[H]
\centering
\setlength{\tabcolsep}{4pt}      
\renewcommand\arraystretch{1.0} 
\begin{tabular}{l | c | cc cc}
\hline
\multirow{2}{*}{\textbf{IoU/Loss}} 
  & \multirow{2}{*}{\textbf{$\mathcal{L}_{\text{ICR}}$}} 
  & \multicolumn{2}{c}{\textbf{AP$^{\text{val}}_{50}$}} 
  & \multicolumn{2}{c}{\textbf{AP$^{\text{val}}_{50:95}$}} \\
  & & A & B & A & B \\
\hline
\multirow{2}{*}{$\mathcal{L}_{\text{IoU}}$}  
  &                  & 65.9 & 97.1 & 38.8 & 81.6 \\
  & \checkmark       & \textbf{68.6} & \textbf{97.4} & \textbf{41.2} & 80.8 \\
\hline
\multirow{2}{*}{$\mathcal{L}_{\text{GIoU}}$} 
  &                  & 62.8 & 97.2 & 36.8 & 81.2 \\
  & \checkmark       & \textbf{67.9} & \textbf{97.4} & \textbf{40.9} & 80.7 \\
\hline
\multirow{2}{*}{$\mathcal{L}_{\text{DIoU}}$} 
  &                  & 62.6 & 97.1 & 37.6 & 80.8 \\
  & \checkmark       & \textbf{68.1} & \textbf{97.2} & \textbf{40.6} & \textbf{81.1} \\
\hline

\multirow{2}{*}{$\mathcal{L}_{\text{CIoU}}$} 
  &                  & 60.6 & 97.5 & 37.3 & 81.4 \\
  & \checkmark       & \textbf{66.7} & 97.4 & \textbf{41.2} & \textbf{81.5} \\
\hline
\end{tabular}
\caption{\textbf{Quantitative comparison of $\mathcal{L}_{\text{ICR}}$ using different based IoU loss} on the SVMLP validation set}
\label{tab:val_ICR_ab}
\end{table}

\subsection{Visualization}

\begin{figure}[!h]
\centering
\includegraphics[width=\columnwidth]{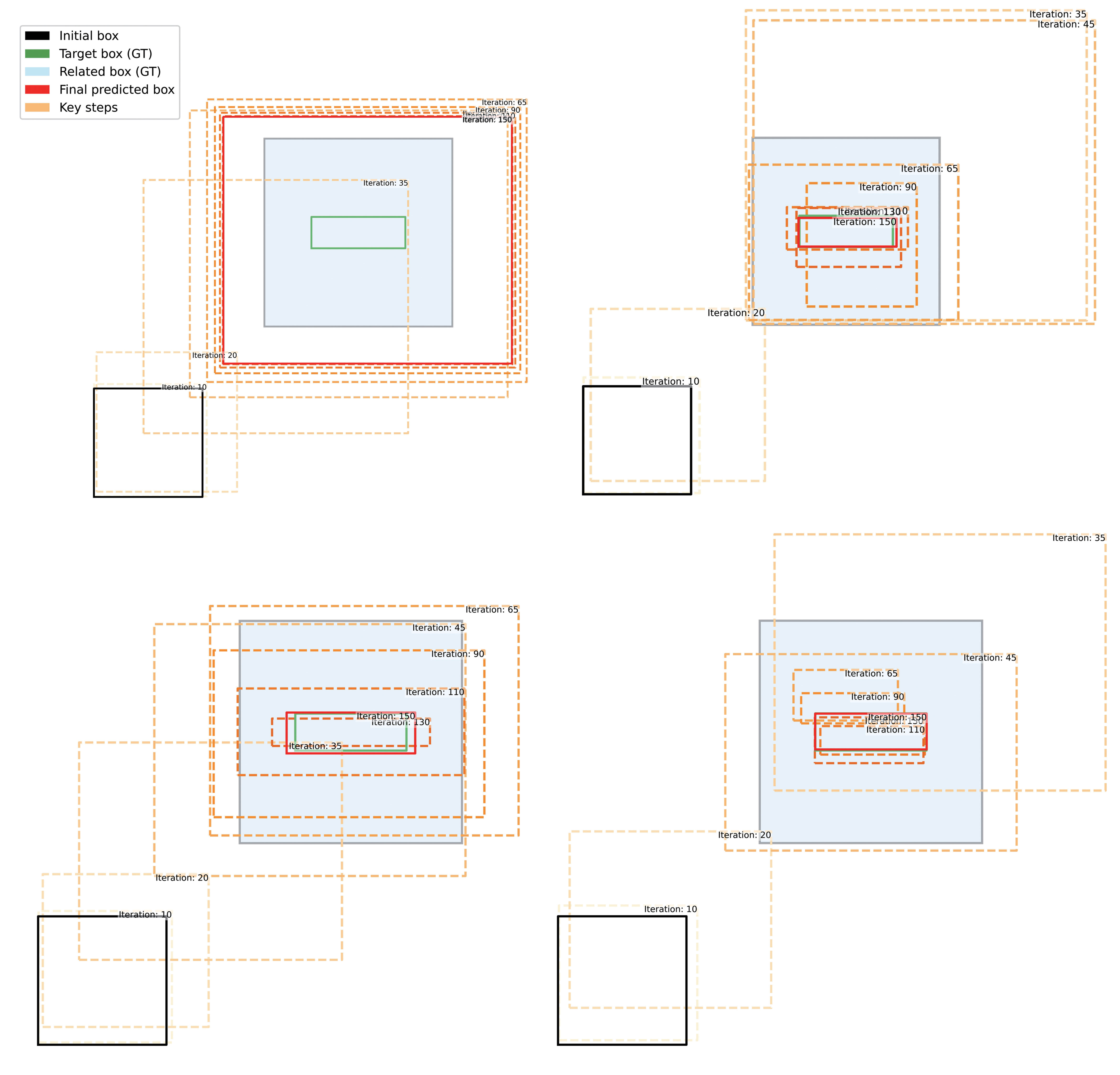}
\caption{
\textbf{Visualization of GIoU (top-left), ICR-GIoU (top-right), CIoU (bottom-left), and ICR-CIoU (bottom-right) losses.} The orange dashed boxes show the key steps in the iterative refinement process and the final prediction in red. ICR-CIoU updates the gradient more quickly and accurately in early training. This also benefits final convergence.
}
\label{fig:ciou_vs_icr}
\end{figure}

The trajectory process is shown in Fig.~\ref{fig:ciou_vs_icr}, the CIoU predicted box smoothly closes to the target green box, then adjusts the width and height to optimize the fit. Our method quickly adjusts the prediction to enter the inter-class bounding box and reduces oscillations while refining the target size.

\begin{figure}[!h]
\centering
\includegraphics[width=\columnwidth]{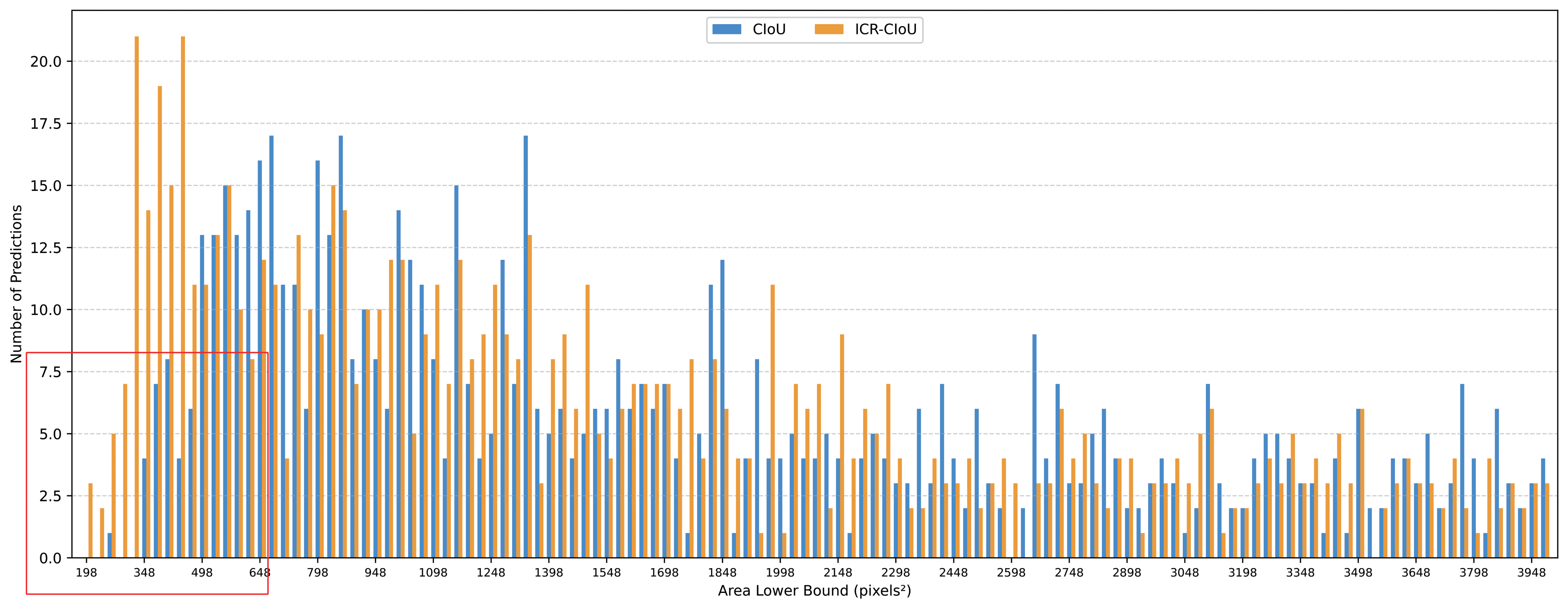}
\caption{
\textbf{Comparing prediction numbers across area intervals} using CIoU and ICR-CIoU on the SVMLP test set using the YOLOV9-T.}
\label{fig:ciou_vs_icr1}
\end{figure}

To demonstrate that ICR loss penalty benefits the deeper feature learning, Fig.~\ref{fig:ciou_vs_icr1} shows the distribution of predicted bounding boxes by area of the objects. Our method shows significant improvement in the area of small objects, ranging from 198 to 498. The minimum detected area is 198 $\mathrm{pixels}^2$, where CIoU is 261 $\mathrm{pixels}^2$.

\section{Conclusion}
The SOD loss problem in MOD is analyzed, and an inter-class loss penalty is proposed, which is available for all IoU-based losses. The more realistic SVMLP dataset is created and used for evaluating the proposed solution. Extensive simulation results show that spatial relationships between small and large classes can benefit small object detection while not harm the learning of other objects. ICR loss penalty provides a simple, efficient way to improve model performance without additional computation. The future work is that various inter-class relations will be researched in MOD.

\bibliographystyle{abbrv}
\bibliography{refs}

\newpage
\section{Appendix}

\subsection{Experimental Settings}
The YOLOv9~\cite{DBLP:conf/eccv/WangYL24} standard Docker is used to set up the Docker environment for the YOLOv9 series. Simply, experiments is implemented in PyTorch 1.14.0 with CUDA 12.0 and conducted on a server with NVIDIA RTX 4090 GPUs. The training hyperparameters are shown in Table~\ref{tab:yolo_hyperparams}.

\begin{table}[h]
\centering
\small
\begin{tabular}{llll}
\toprule
\textbf{Parameter} & \textbf{Value} & \textbf{Parameter} & \textbf{Value}\\
\midrule
\texttt{lr0} & 0.01 & \texttt{momentum} & 0.937 \\
\texttt{lrf} & 0.01 & \texttt{weight\_decay} & 0.0005 \\
\texttt{box} & 7.5 & \texttt{warmup\_epochs} & 3.0 \\
\texttt{cls} & 0.5 & \texttt{warmup\_momentum} & 0.8 \\
\texttt{cls\_pw} & 1.0 & \texttt{warmup\_bias\_lr} & 0.1 \\
\texttt{obj} & 0.7 & \texttt{obj\_pw} & 1.0 \\
\texttt{dfl} & 1.5 & \texttt{anchor\_t} & 5.0 \\
\texttt{iou\_t} & 0.20 & \texttt{fl\_gamma} & 0.0 \\
\texttt{hsv\_h} & 0.015 & \texttt{translate} & 0.1 \\
\texttt{hsv\_s} & 0.7 & \texttt{scale} & 0.9 \\
\texttt{shear} & 0.0 & \texttt{perspective} & 0.0 \\
\texttt{degrees} & 0.0 & \texttt{flipud} & 0.0 \\
\texttt{hsv\_v} & 0.47 & \texttt{fliplr} & 0.5 \\
\texttt{mixup} & 0.15 & \texttt{mosaic} & 1.0 \\
\texttt{copy\_paste} & 0.3 \\
\bottomrule
\end{tabular}
\caption{YOLOv9 series training hyperparameters and data augmentation configuration used in our experiments.}
\label{tab:yolo_hyperparams}
\end{table}

The Ultralytics standard Docker is used to set up the Docker environment for YOLOv11~\cite{DBLP:journals/corr/abs-2410-17725}, YOLOv12~\cite{DBLP:journals/corr/abs-2502-12524} and UAV-DETR~\cite{DBLP:journals/corr/abs-2501-01855}, which are set up in PyTorch 2.7.0 with CUDA 12.6. The default setting of their training hyperparameters is used as provided by Ultralytics.

For RE-DETRv2~\cite{DBLP:journals/corr/abs-2407-17140}, the open source codes from the official RT-DETRv2 repository and the same Docker environment as YOLOV11 are used. The loss weight setting is followed by RE-DETR~\cite{DBLP:conf/cvpr/ZhaoLXWWDLC24}. According to SVMLP dataset, the data argumentation configuration for 50 epochs is shown in Table~\ref{tab:rtdetr_hyperparams}.

\begin{table}[!h]
\centering
\small
\begin{tabular}{ll}
\toprule
\textbf{Parameter} & \textbf{Value} \\
\midrule
\texttt{RandomPhotometricDistort} & 0.5 \\
\texttt{RandomZoomOut} & - \\
\texttt{RandomIoUCrop} & 0.47 \\
\texttt{SanitizeBoundingBoxes} & 1 \\
\texttt{RandomHorizontalFlip} & - \\
\bottomrule
\end{tabular}
\caption{RT-DETRv2 data augmentation configuration used in experiments based on SVMLP.}
\label{tab:rtdetr_hyperparams}
\end{table}

\subsection{Visualization}
\begin{figure}[h]
    \captionsetup{skip=2pt}
    \setlength{\abovecaptionskip}{2pt}
    \setlength{\belowcaptionskip}{-2pt}

    \centering
    \setlength{\tabcolsep}{0pt}
    \renewcommand{\arraystretch}{0.9}

    \begin{tabular}{@{}c@{\hspace{1pt}}c@{}}
        \includegraphics[width=0.47\columnwidth]{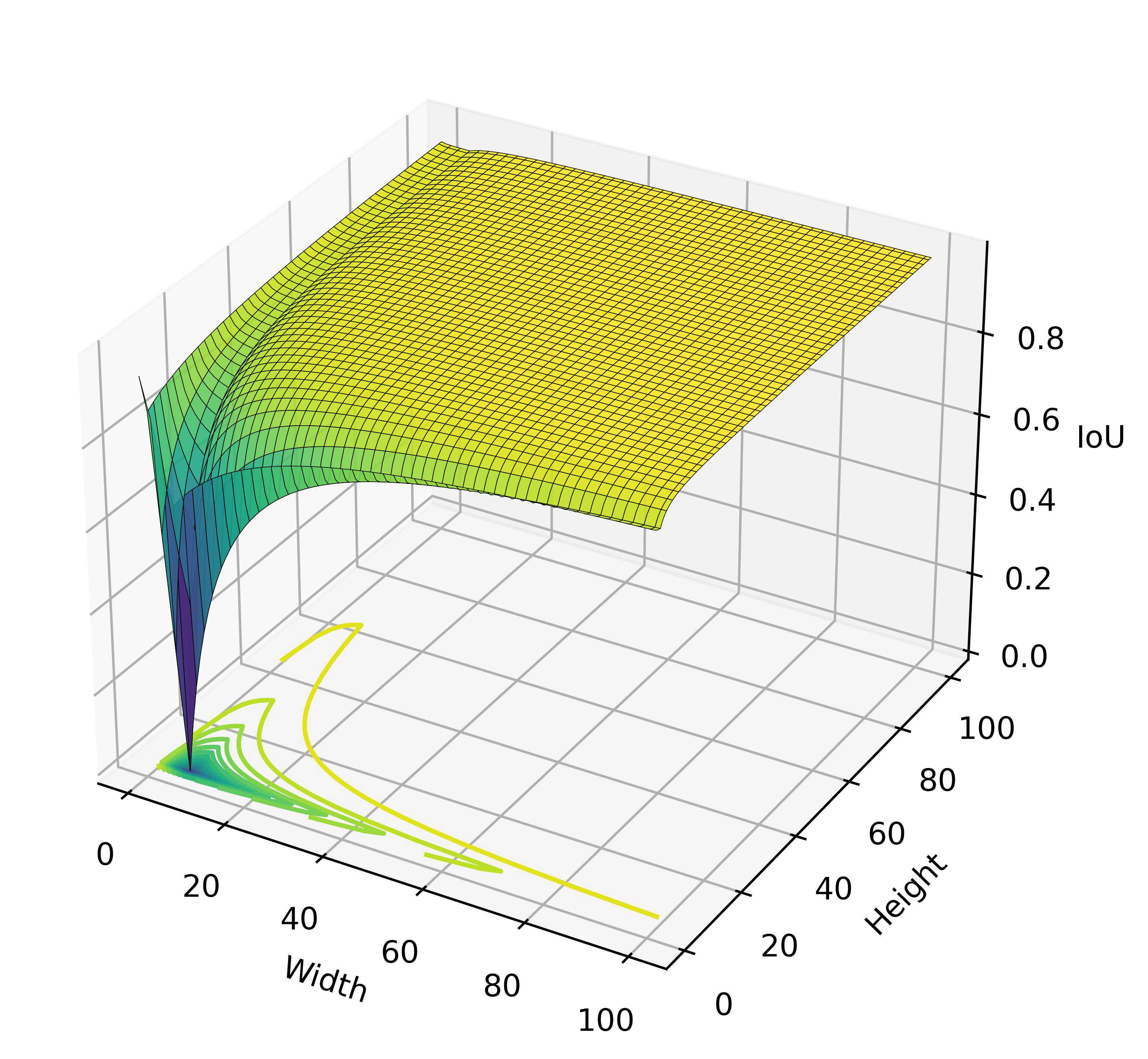} &
        \includegraphics[width=0.47\columnwidth]{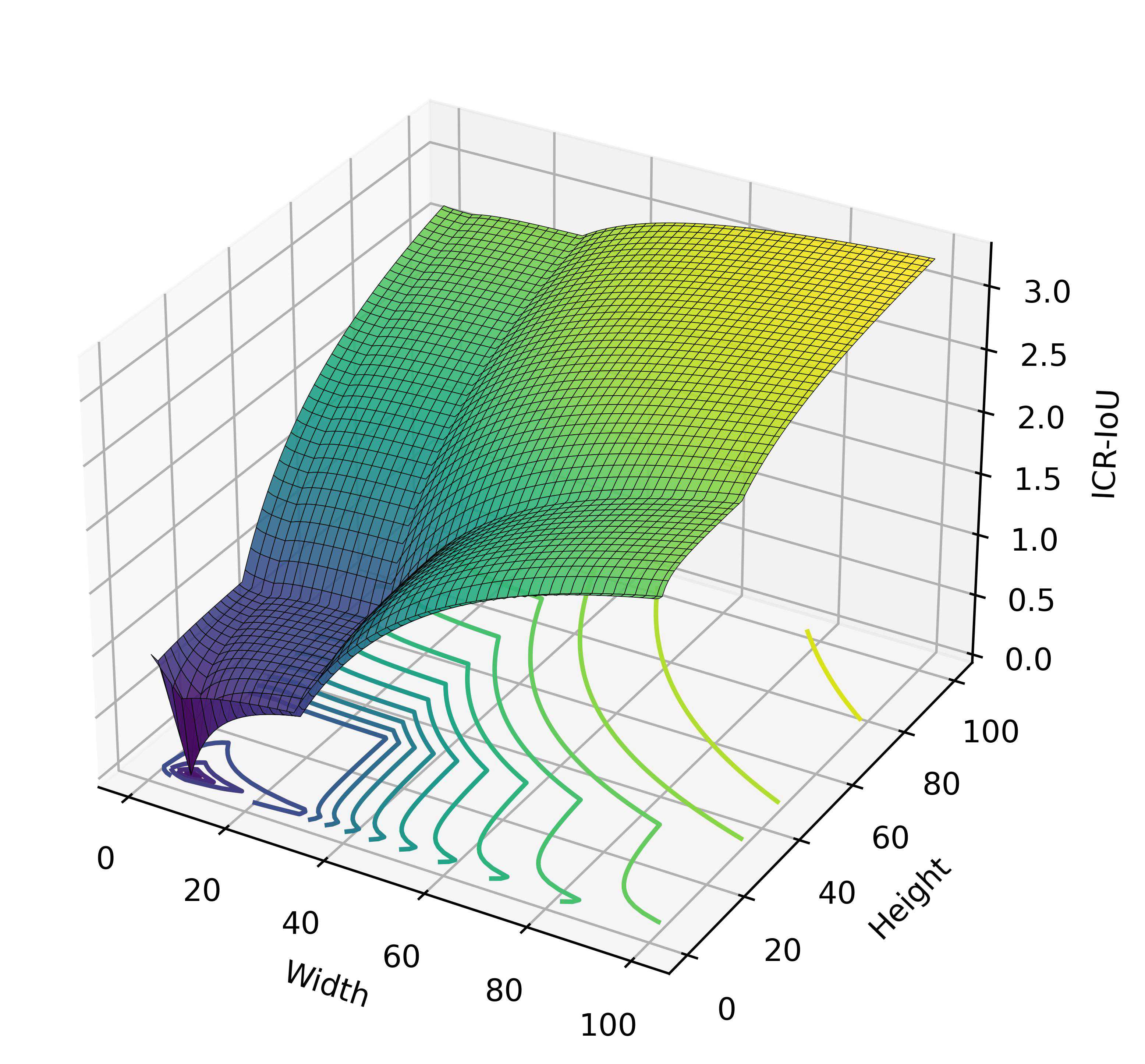} \\
        {\scriptsize IoU} & {\scriptsize ICR-IoU} \\[-2pt]

        \includegraphics[width=0.47\columnwidth]{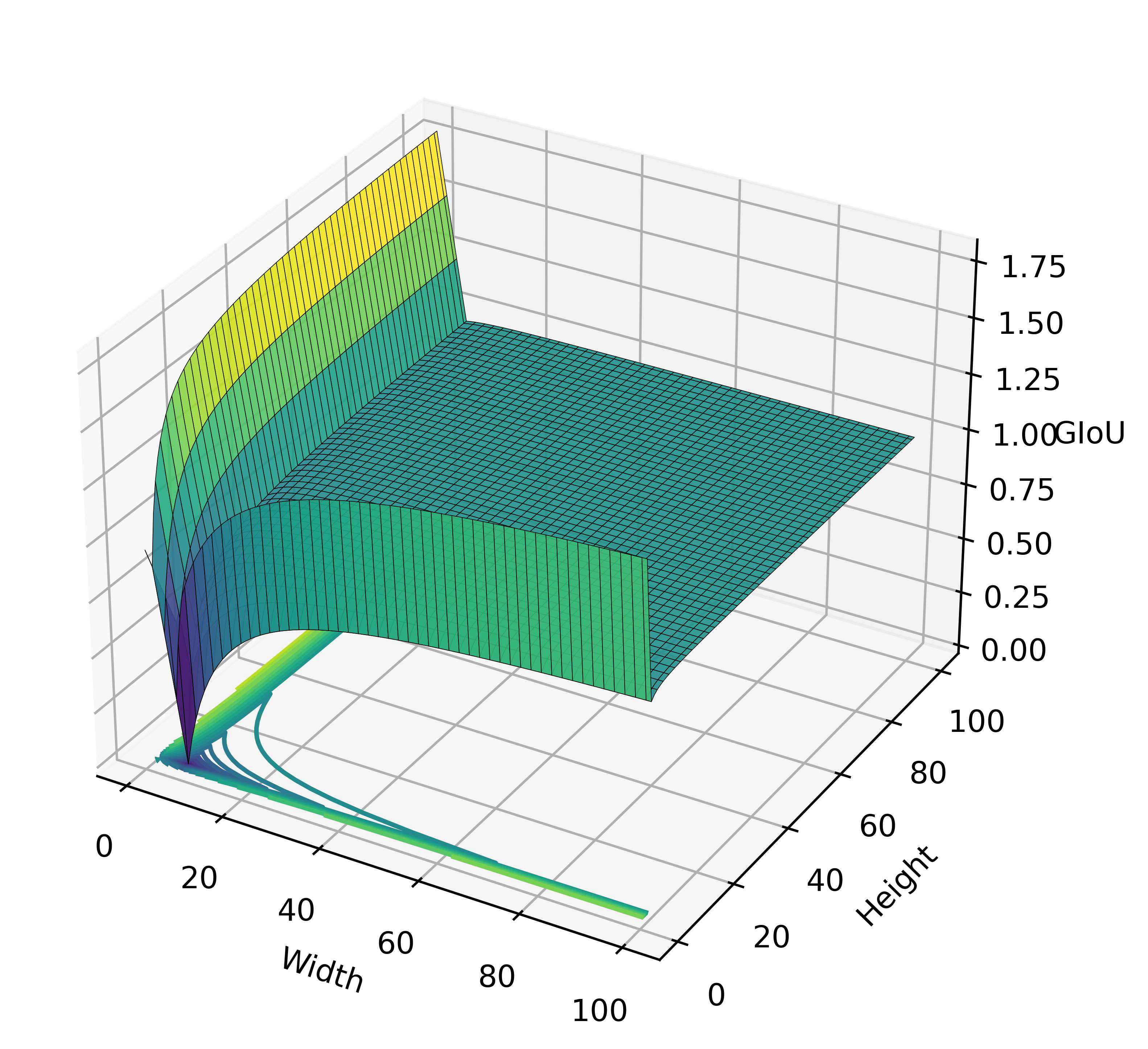} &
        \includegraphics[width=0.47\columnwidth]{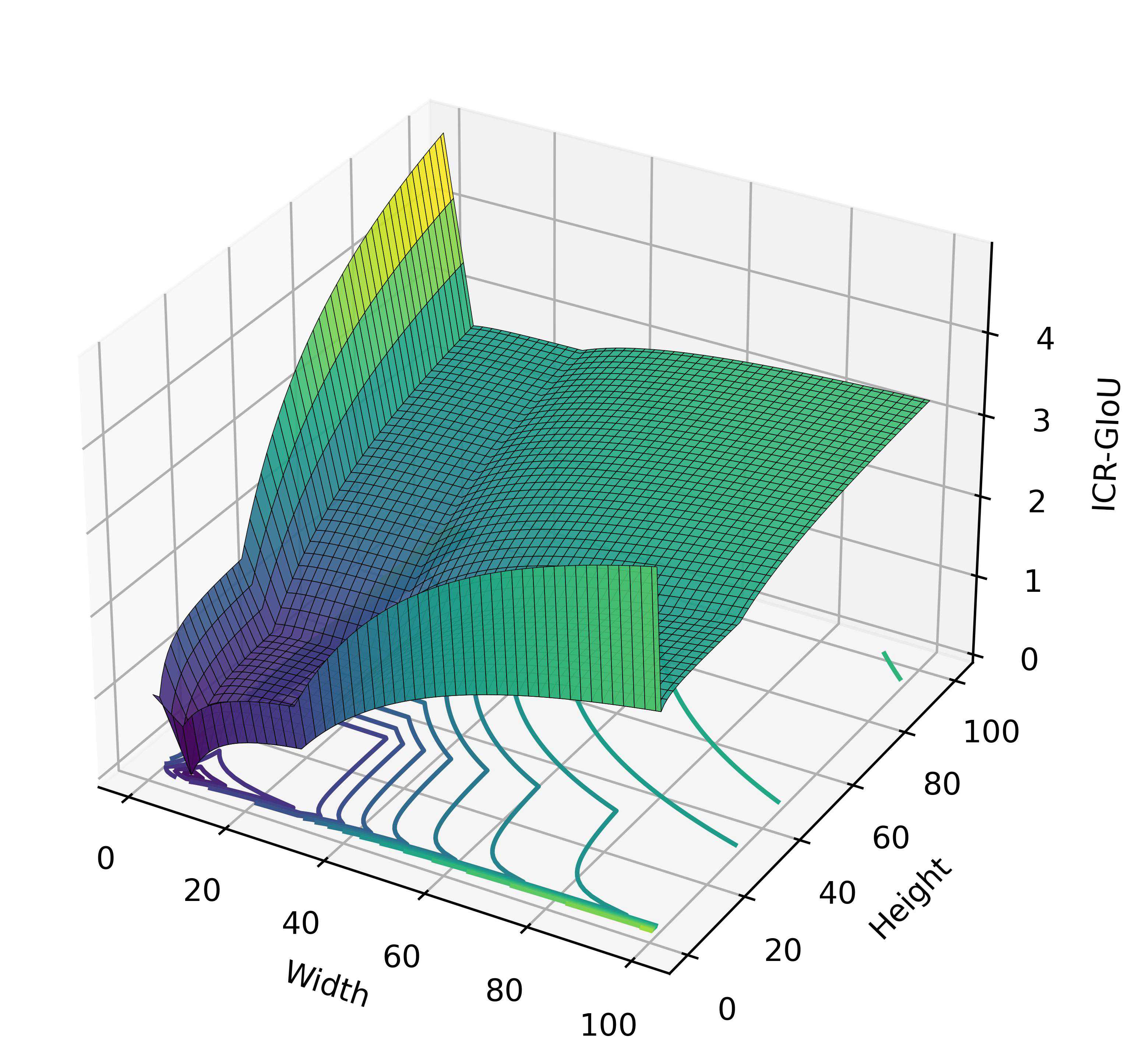} \\
        {\scriptsize GIoU} & {\scriptsize ICR-GIoU} \\[-2pt]

        \includegraphics[width=0.47\columnwidth]{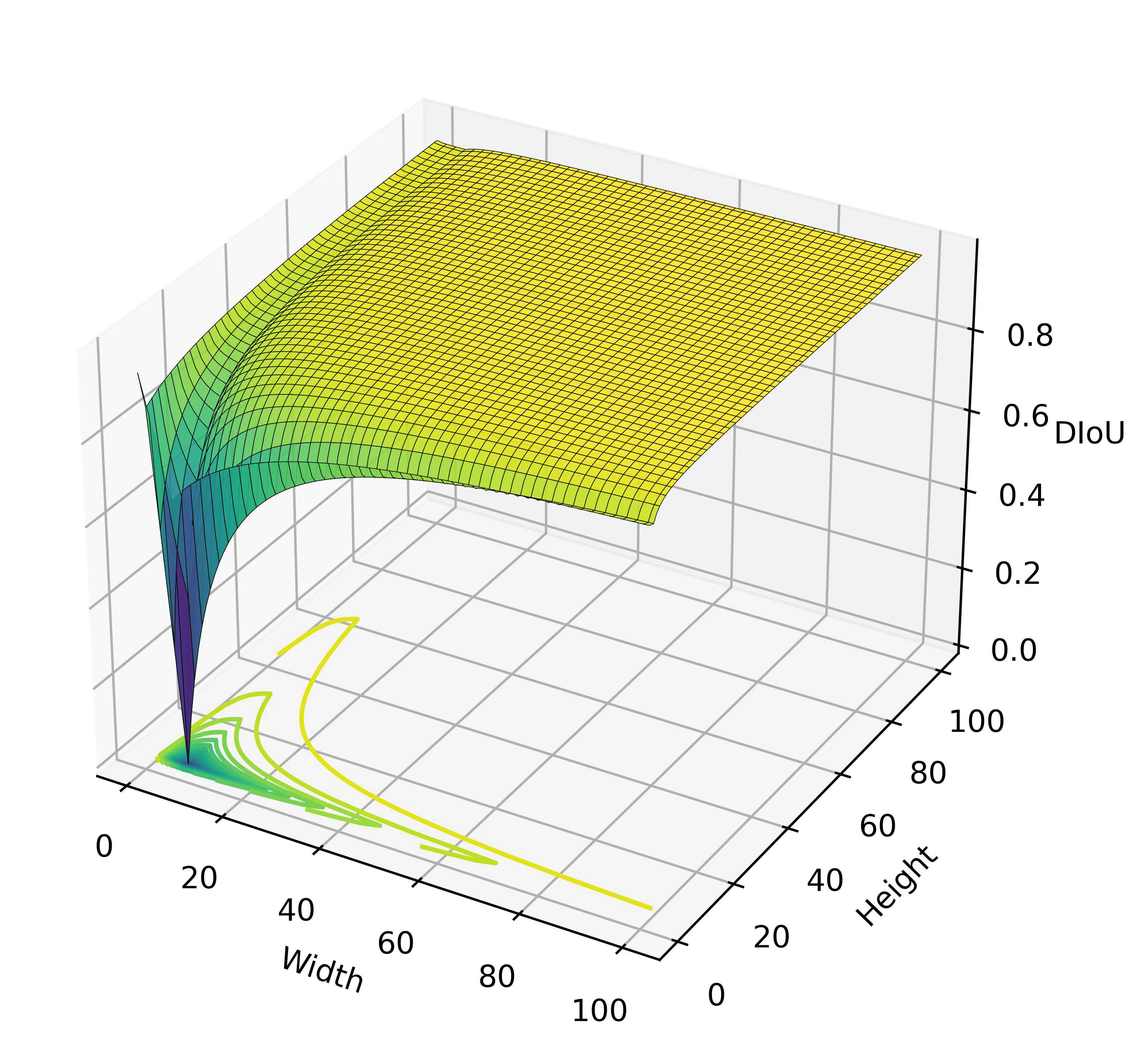} &
        \includegraphics[width=0.47\columnwidth]{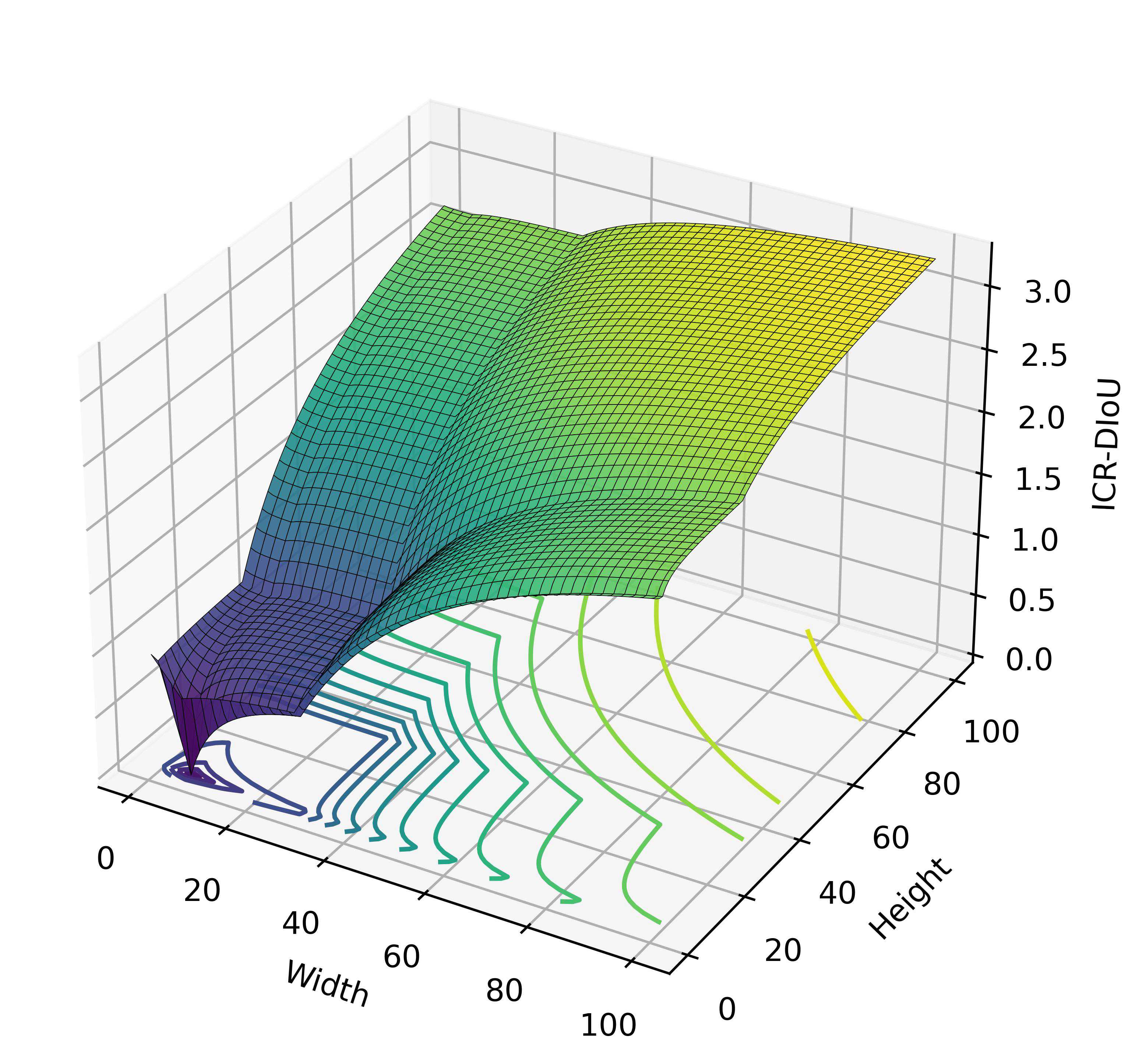} \\
        {\scriptsize DIoU} & {\scriptsize ICR-DIoU} \\[-2pt]

        \includegraphics[width=0.47\columnwidth]{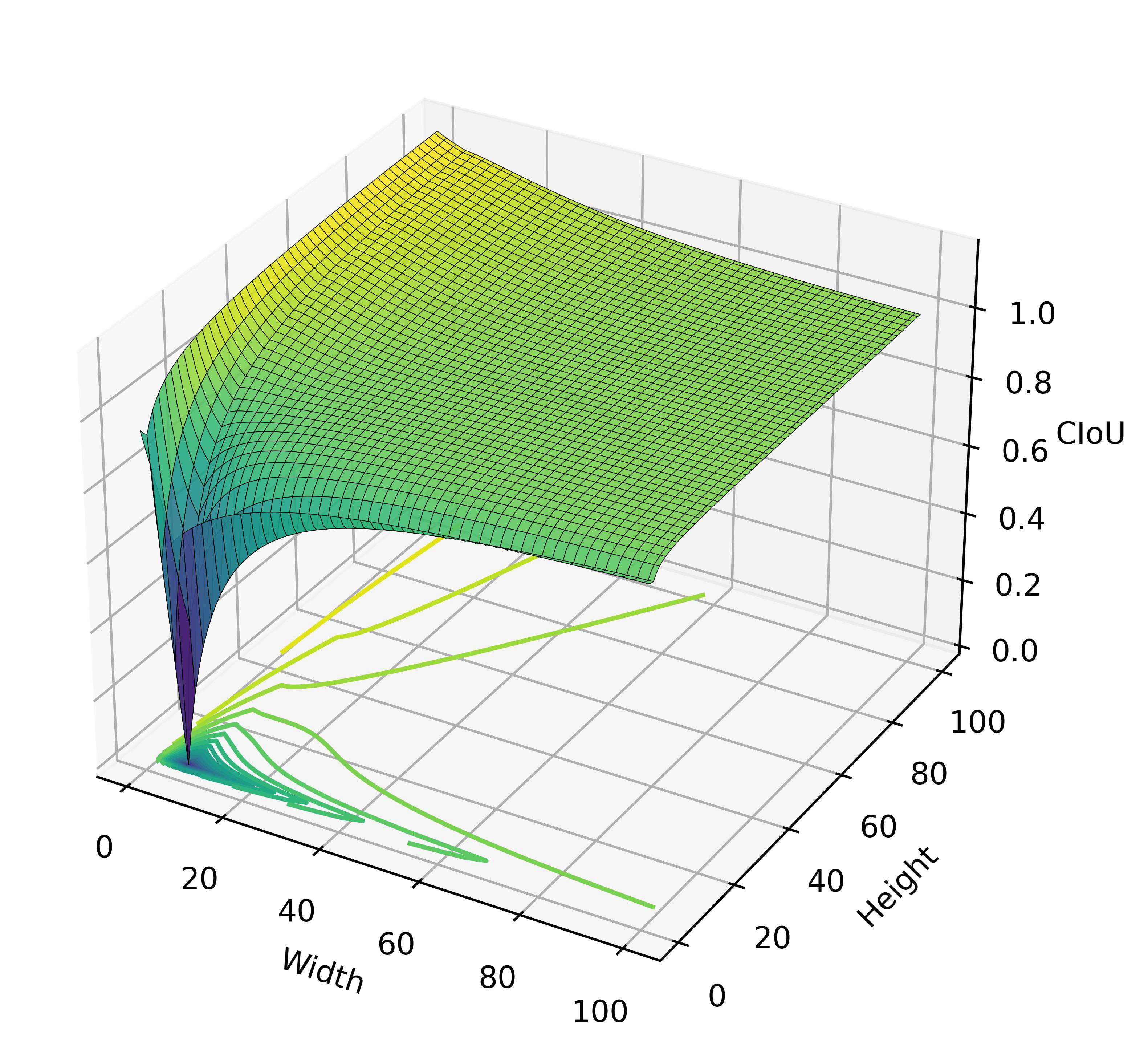} &
        \includegraphics[width=0.47\columnwidth]{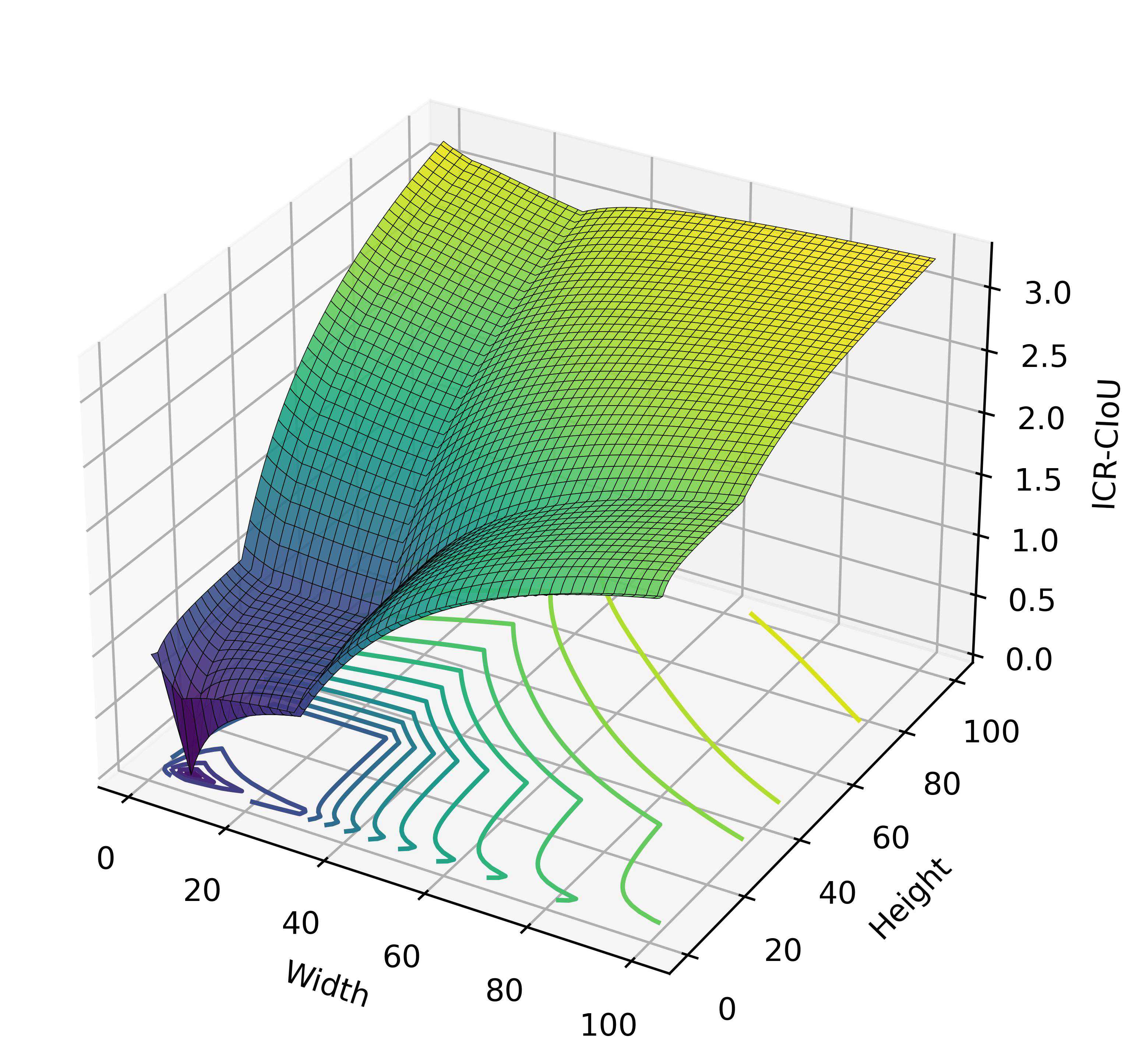} \\
        {\scriptsize CIoU} & {\scriptsize ICR-CIoU}
    \end{tabular}

    \caption{Comparison of the IoU-based losses (left) and their ICR variants (right).}
    \label{fig:all_losses}
\end{figure}

To better show the difference in the gradient, the extended figures for ICR-IoU, ICR-GIoU, and ICR-DIoU are given in Fig.~\ref{fig:all_losses}. The proposed ICR loss penalty works across different IoU-based algorithms and updates a better gradient, especially when the predicted bounding box is far from the targets.

\begin{figure*}[!h]
\centering

\begin{subfigure}{0.47\textwidth}
    \includegraphics[width=\linewidth]{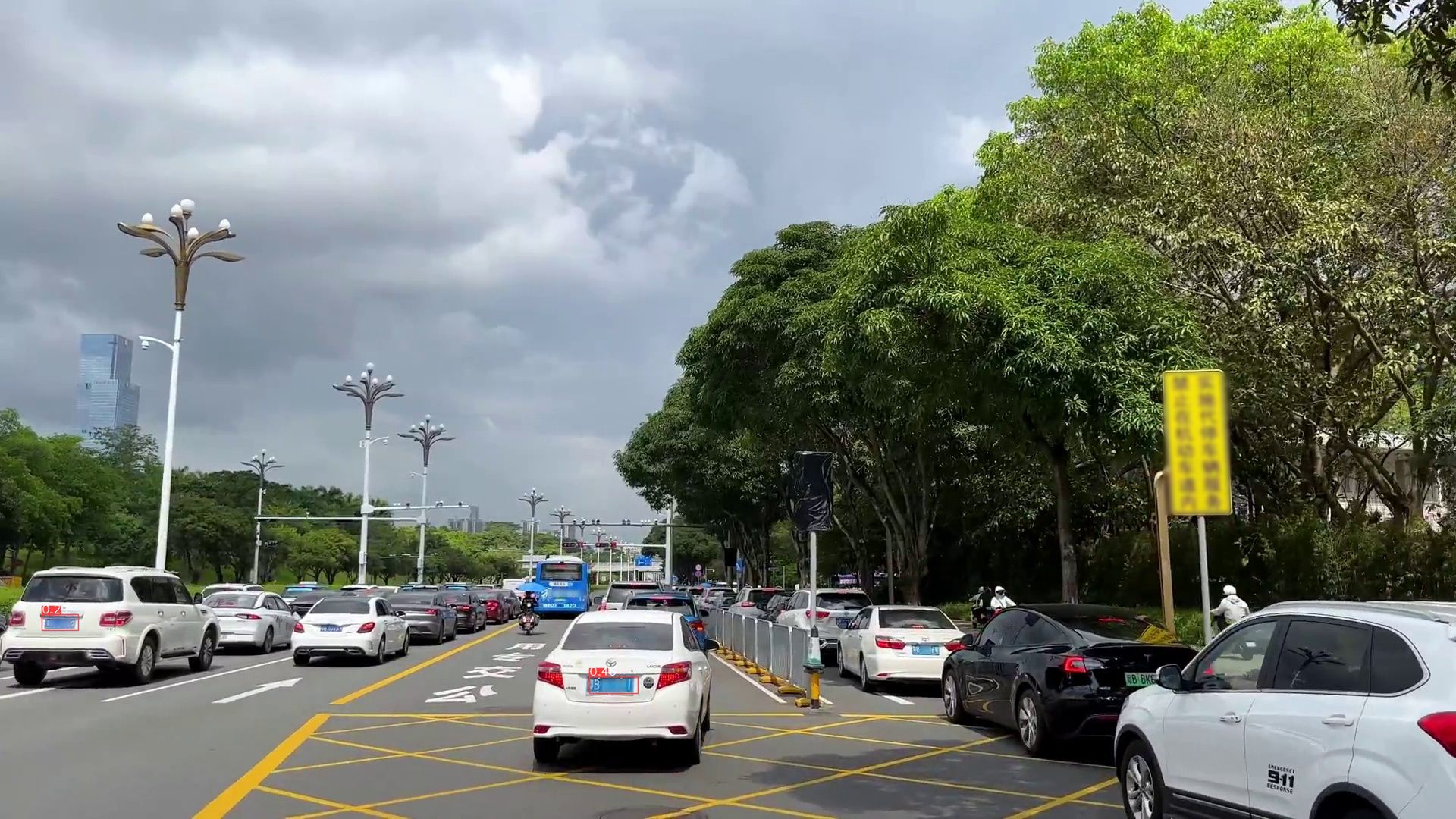}
\end{subfigure}
\hspace{0.02\textwidth}
\begin{subfigure}{0.47\textwidth}
    \includegraphics[width=\linewidth]{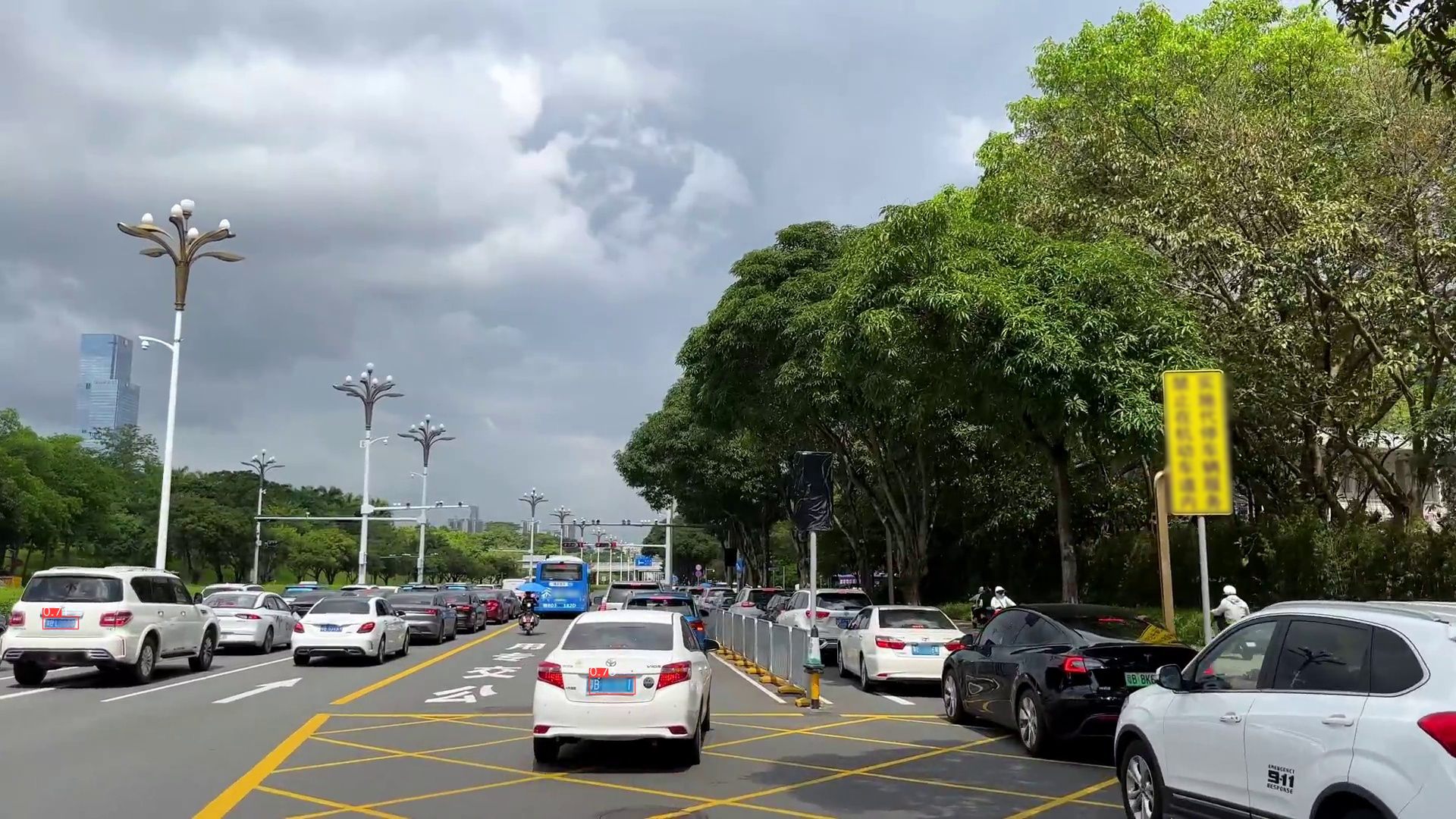}
\end{subfigure}

\vspace{1em}

\begin{subfigure}{0.47\textwidth}
    \includegraphics[width=\linewidth]{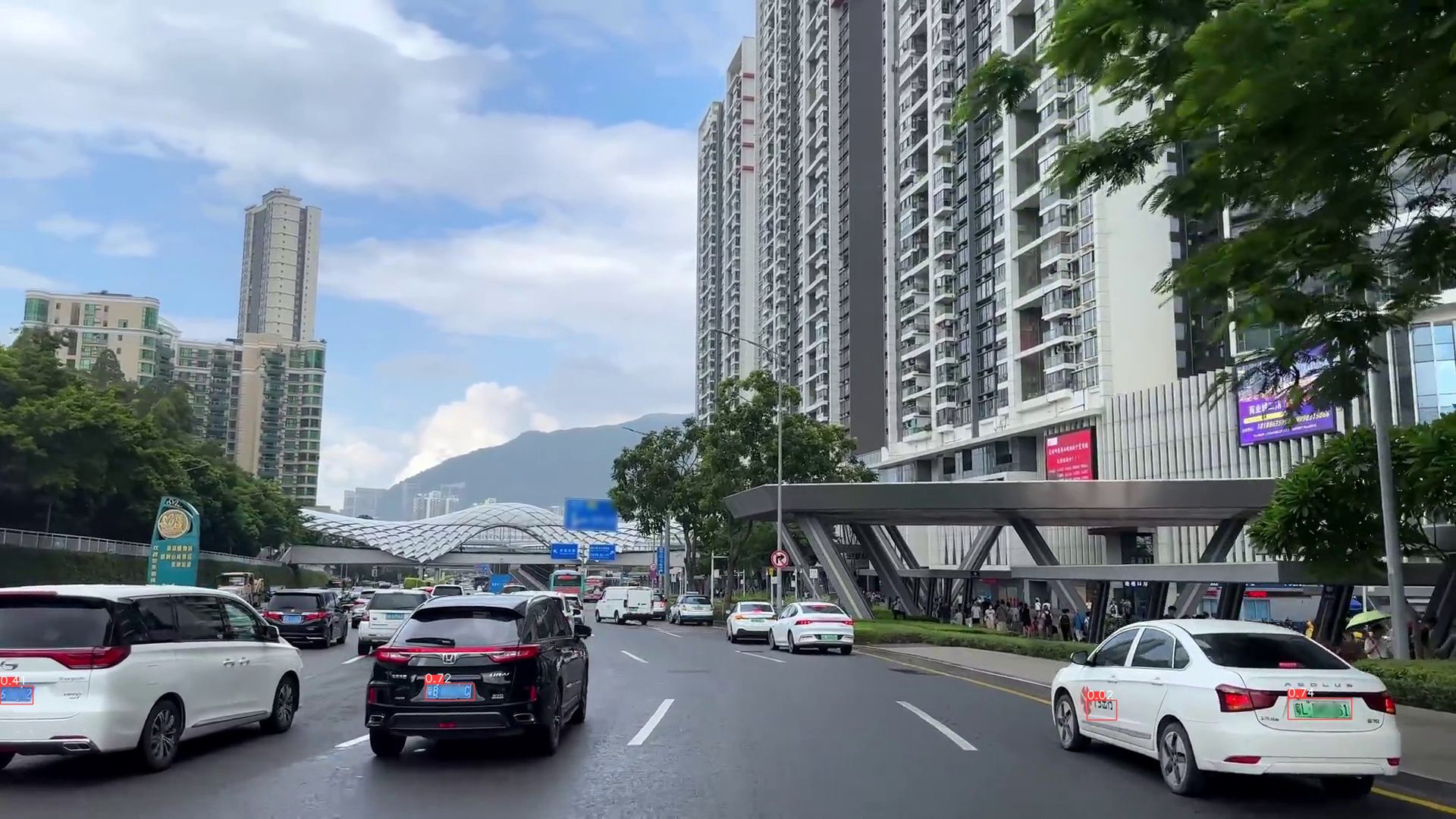}
\end{subfigure}
\hspace{0.02\textwidth}
\begin{subfigure}{0.47\textwidth}
    \includegraphics[width=\linewidth]{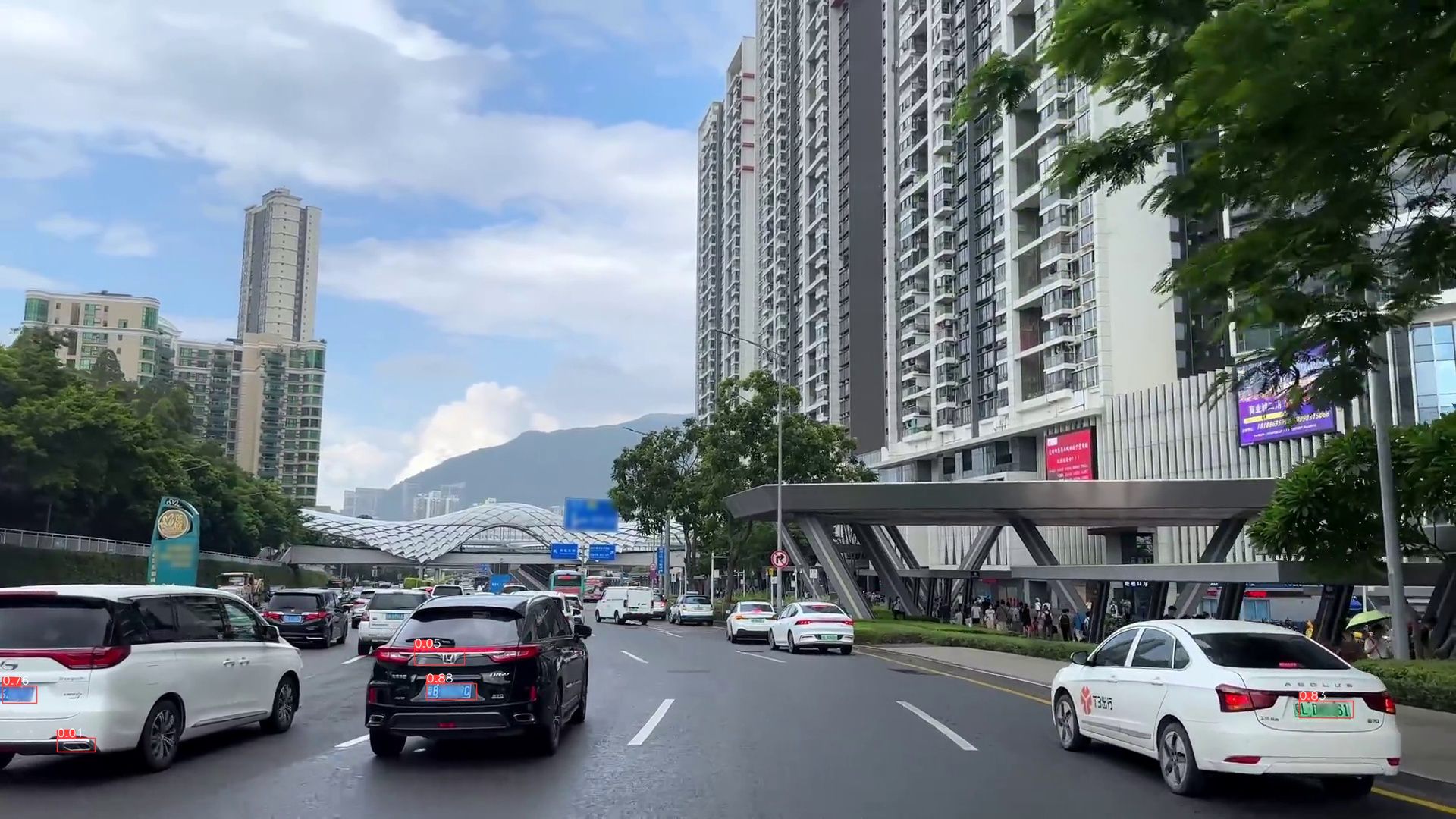}
\end{subfigure}

\vspace{1em}

\begin{subfigure}{0.47\textwidth}
    \includegraphics[width=\linewidth]{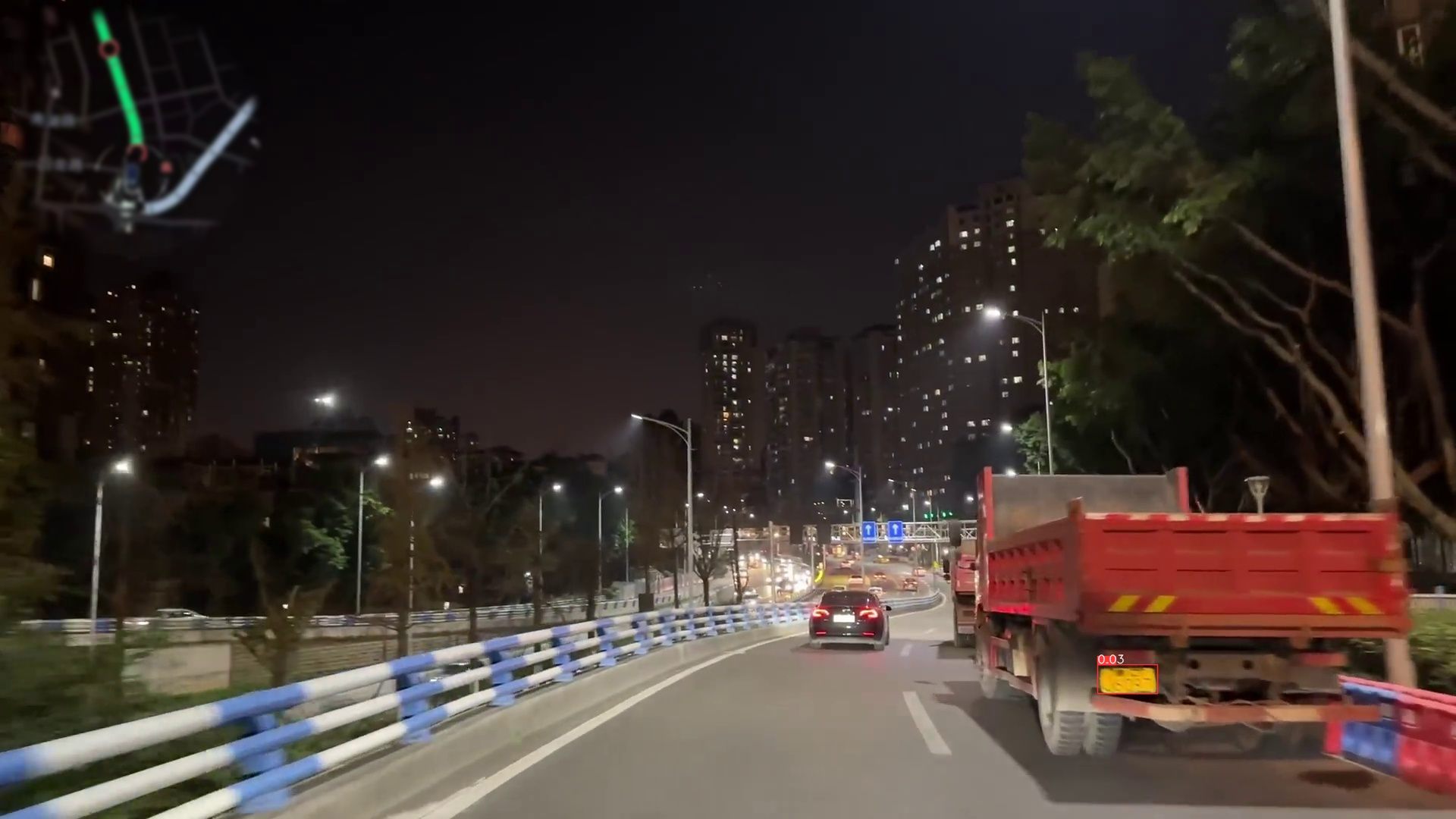}
\end{subfigure}
\hspace{0.02\textwidth}
\begin{subfigure}{0.47\textwidth}
    \includegraphics[width=\linewidth]{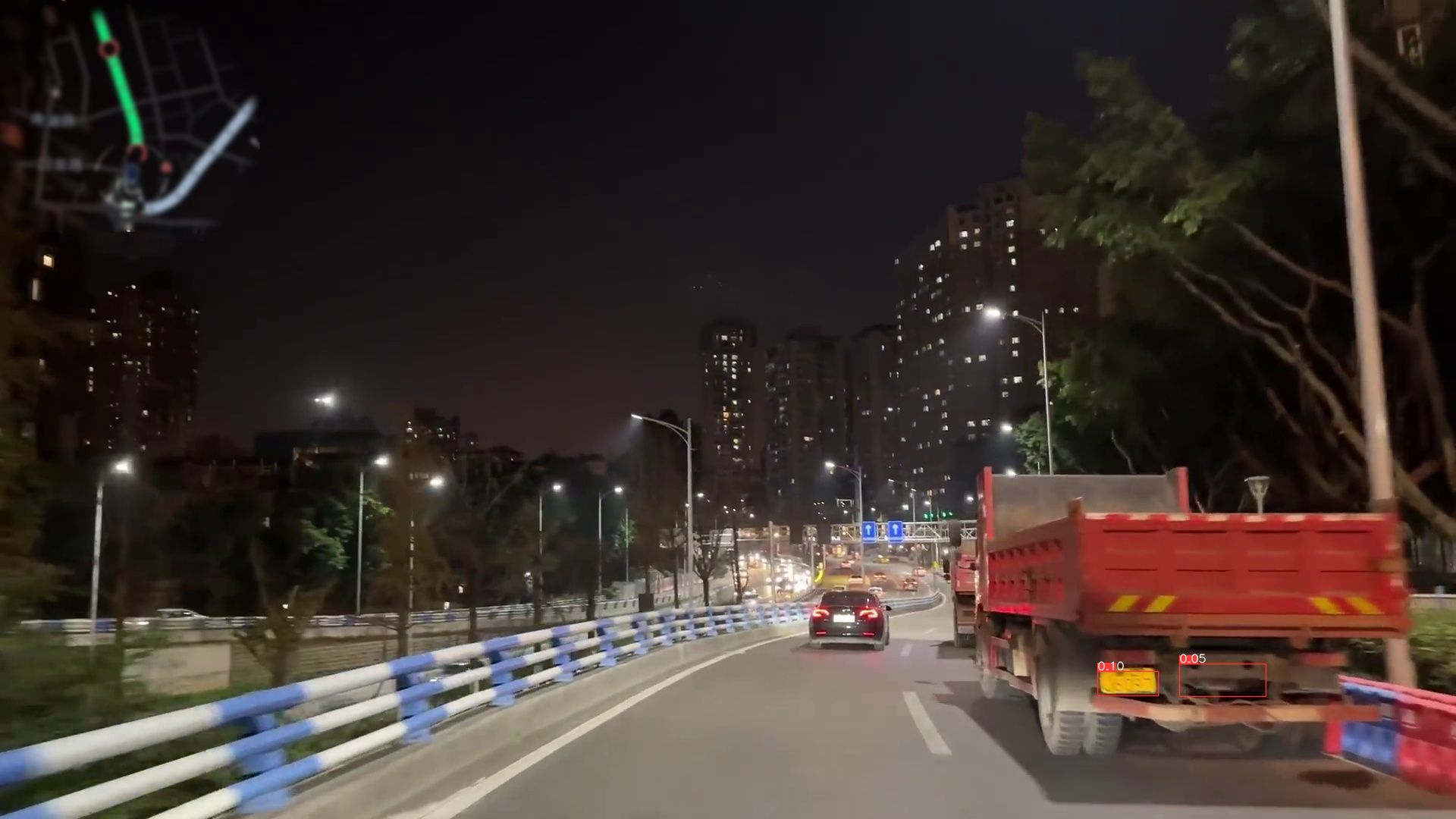}
\end{subfigure}

\vspace{1em}

\begin{subfigure}{0.47\textwidth}
    \includegraphics[width=\linewidth]{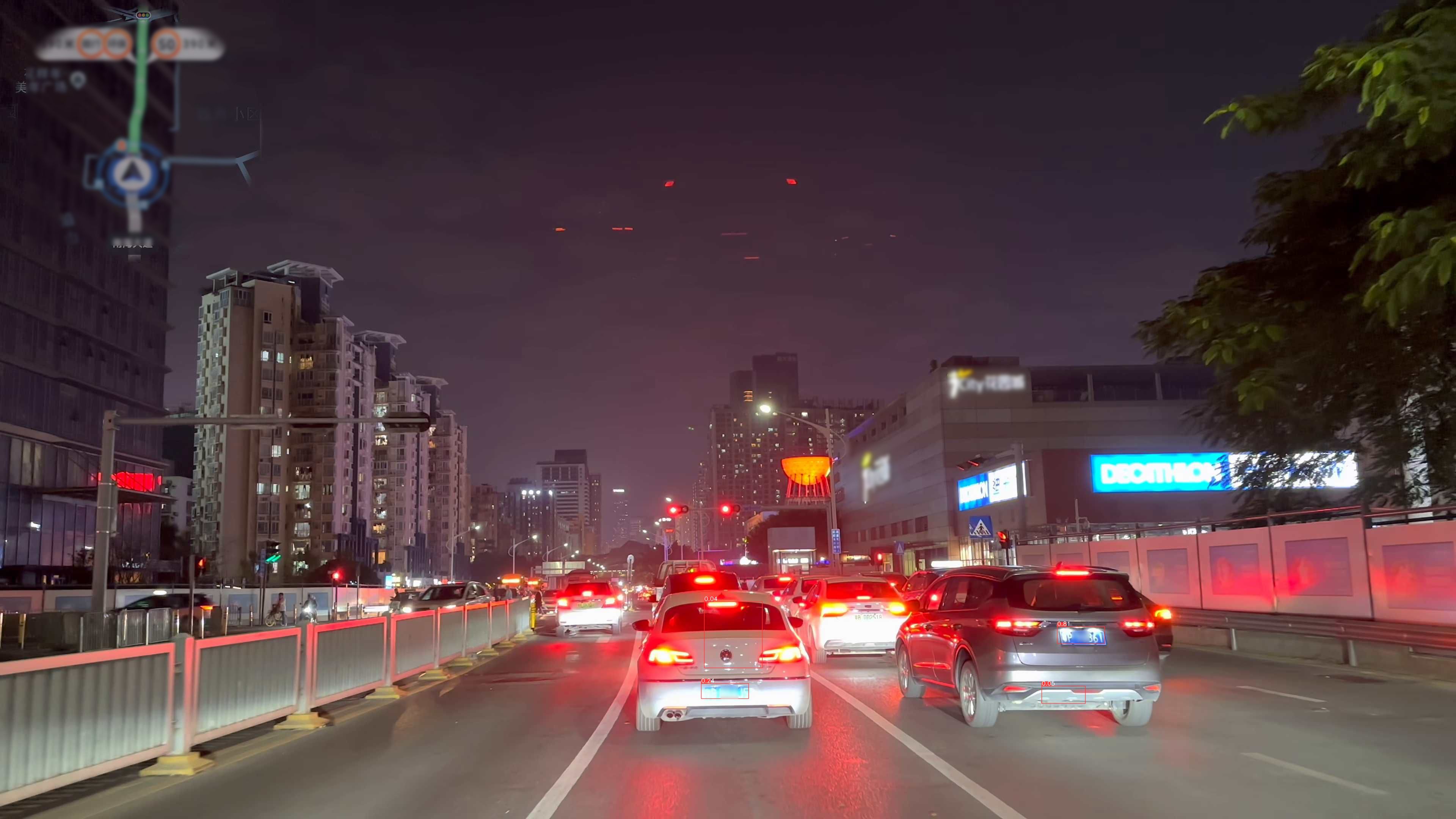}
\end{subfigure}
\hspace{0.02\textwidth}
\begin{subfigure}{0.47\textwidth}
    \includegraphics[width=\linewidth]{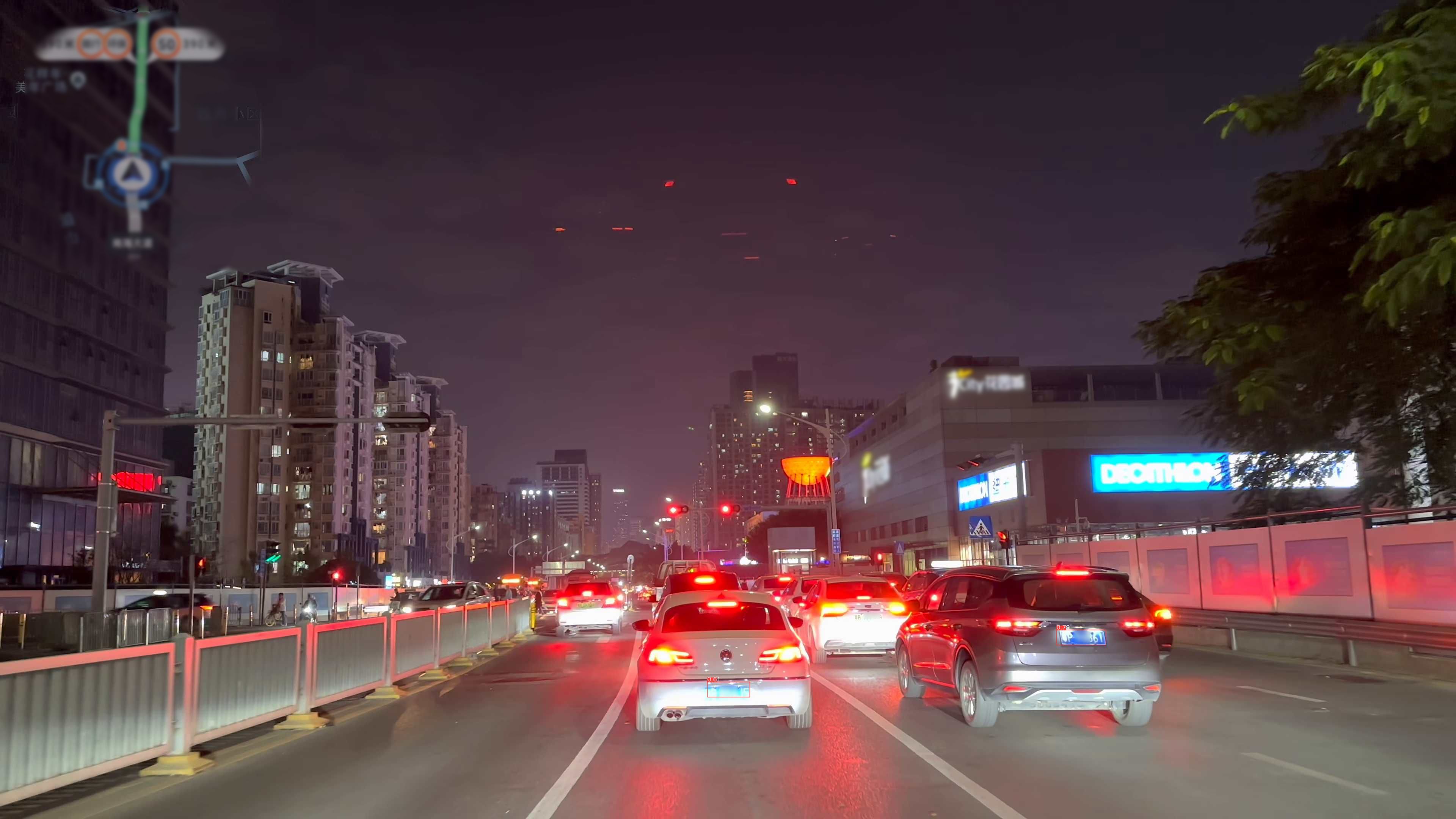}
\end{subfigure}

\caption{
Visualization of YOLOV9-T at Epoch 20 for four different images. Each row shows the CIoU loss (left) and the ICR-CIoU loss prediction (right) on SVMLP dataset.
}
\label{fig:epoch20_large}
\end{figure*}

\begin{figure*}[h]
\centering

\begin{subfigure}{0.47\textwidth}
    \includegraphics[width=\linewidth]{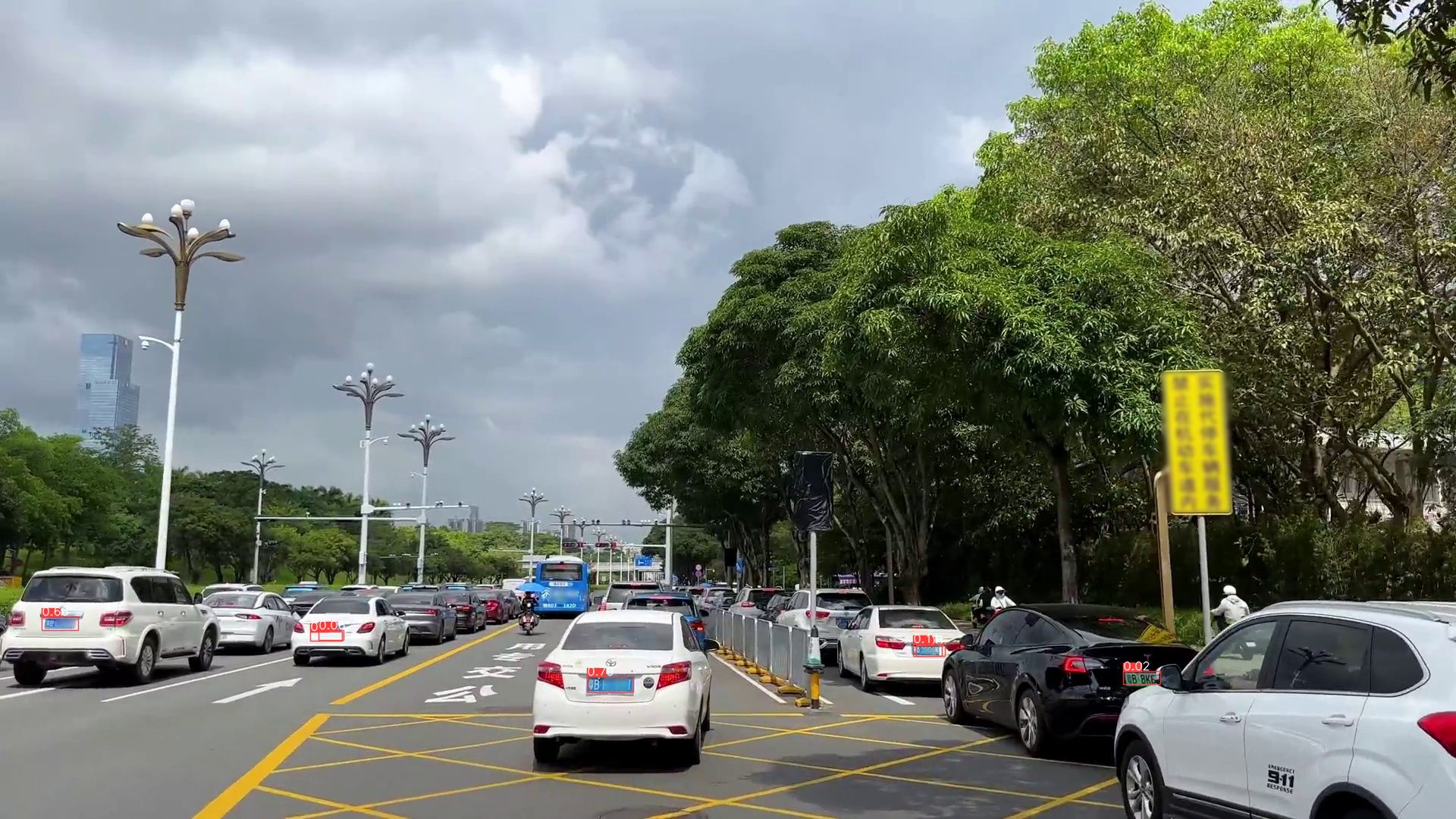}
\end{subfigure}
\hspace{0.02\textwidth}
\begin{subfigure}{0.47\textwidth}
    \includegraphics[width=\linewidth]{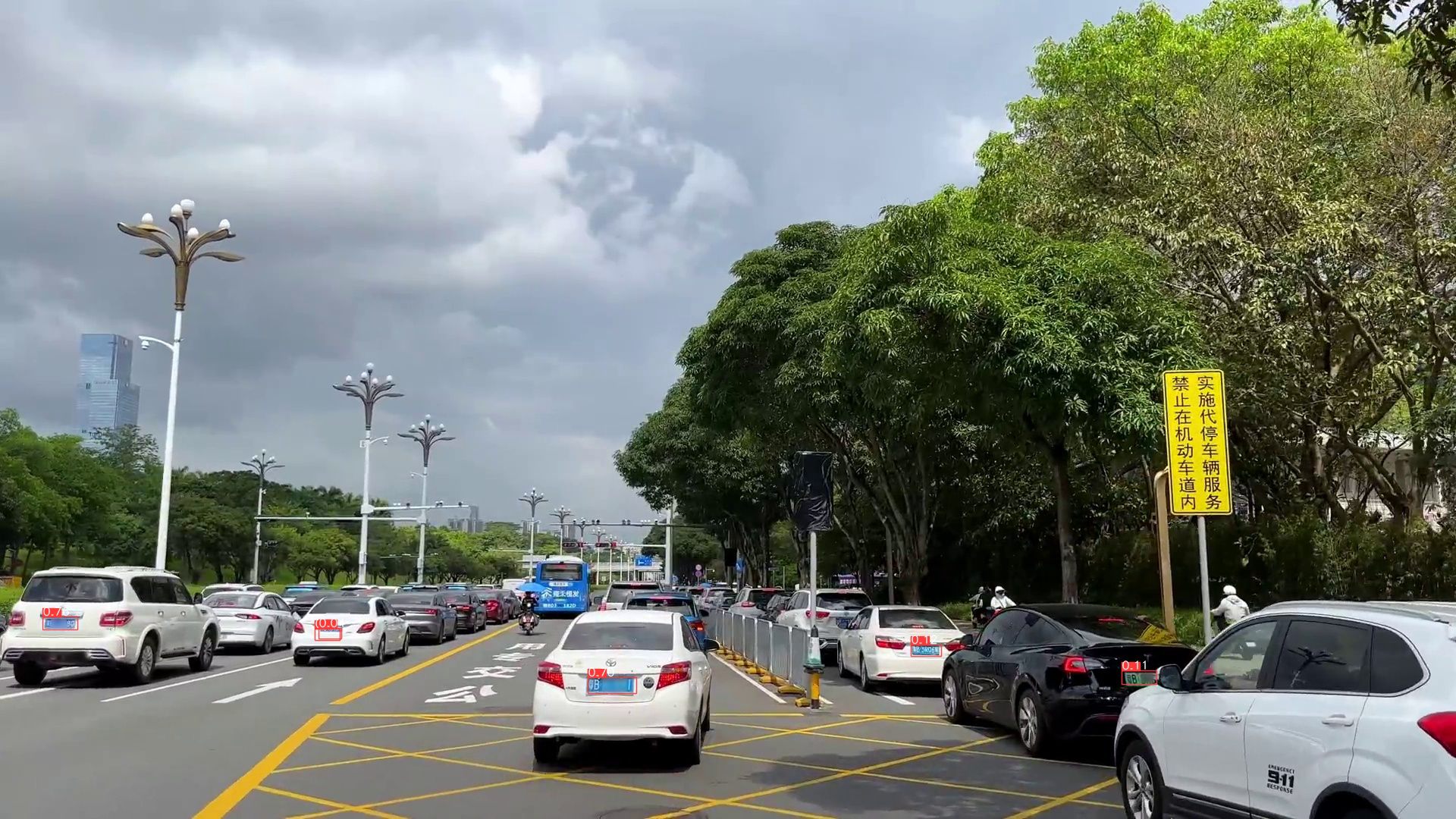}
\end{subfigure}

\vspace{1em}

\begin{subfigure}{0.47\textwidth}
    \includegraphics[width=\linewidth]{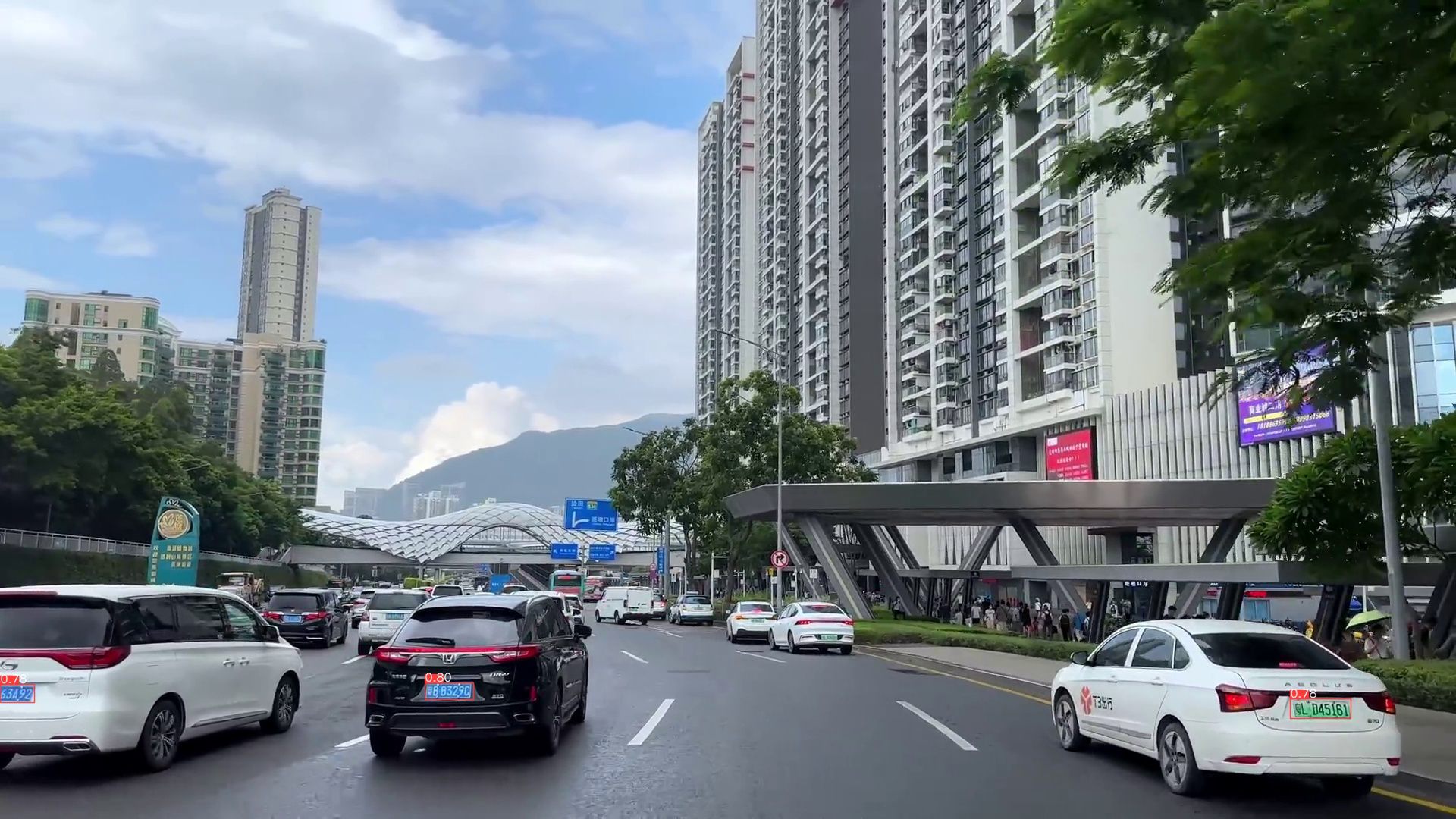}
\end{subfigure}
\hspace{0.02\textwidth}
\begin{subfigure}{0.47\textwidth}
    \includegraphics[width=\linewidth]{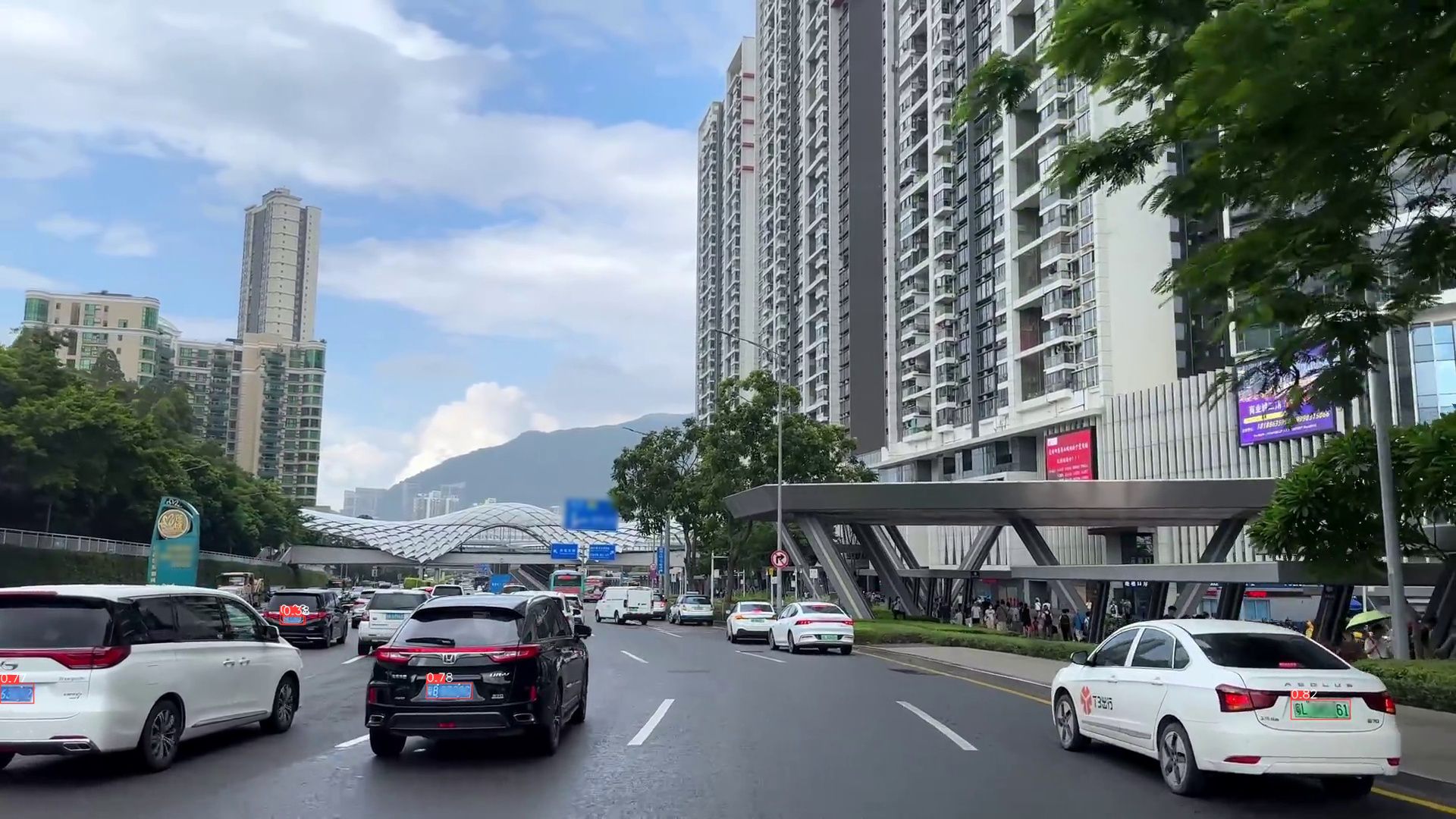}
\end{subfigure}

\vspace{1em}

\begin{subfigure}{0.47\textwidth}
    \includegraphics[width=\linewidth]{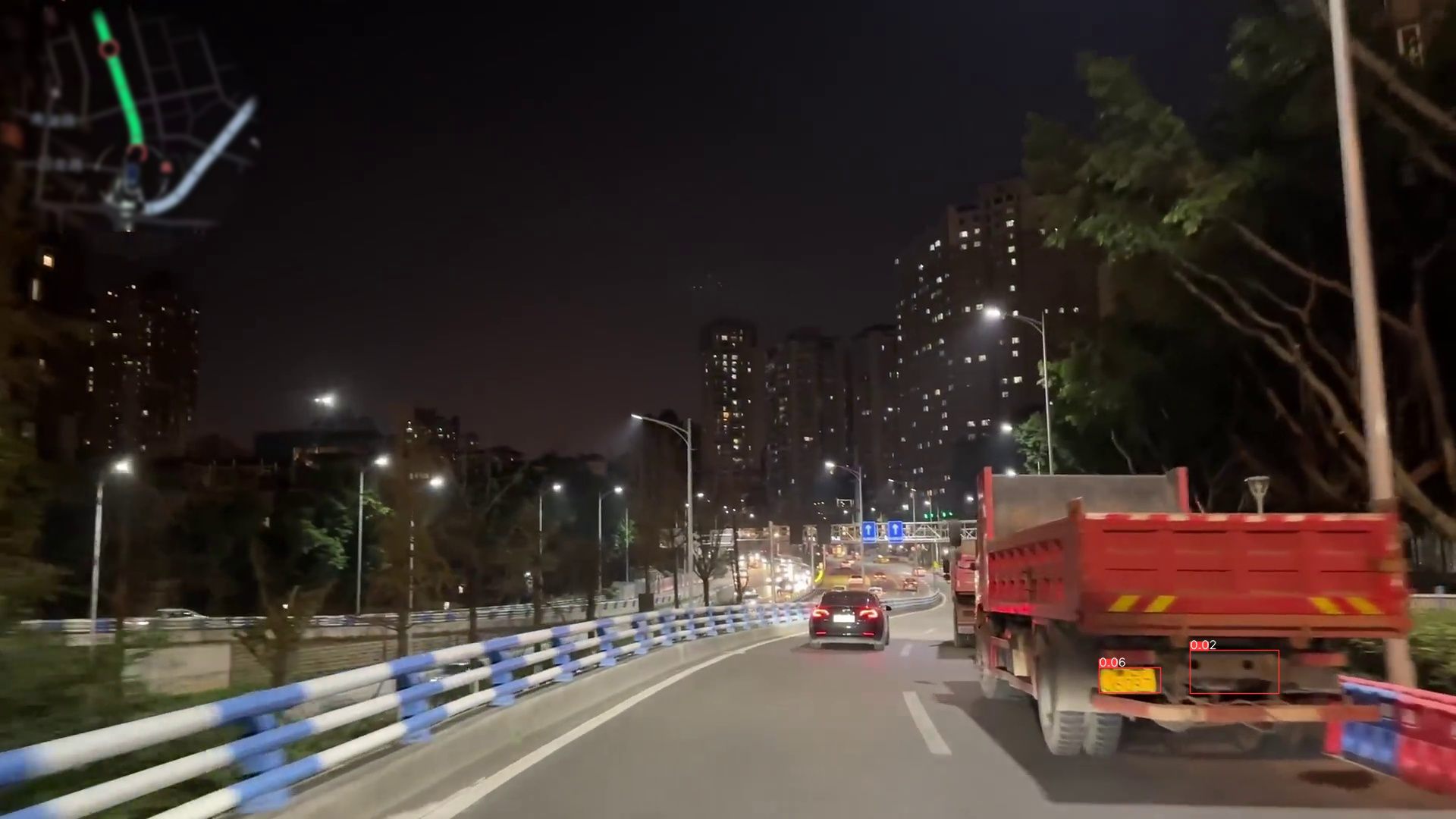}
\end{subfigure}
\hspace{0.02\textwidth}
\begin{subfigure}{0.47\textwidth}
    \includegraphics[width=\linewidth]{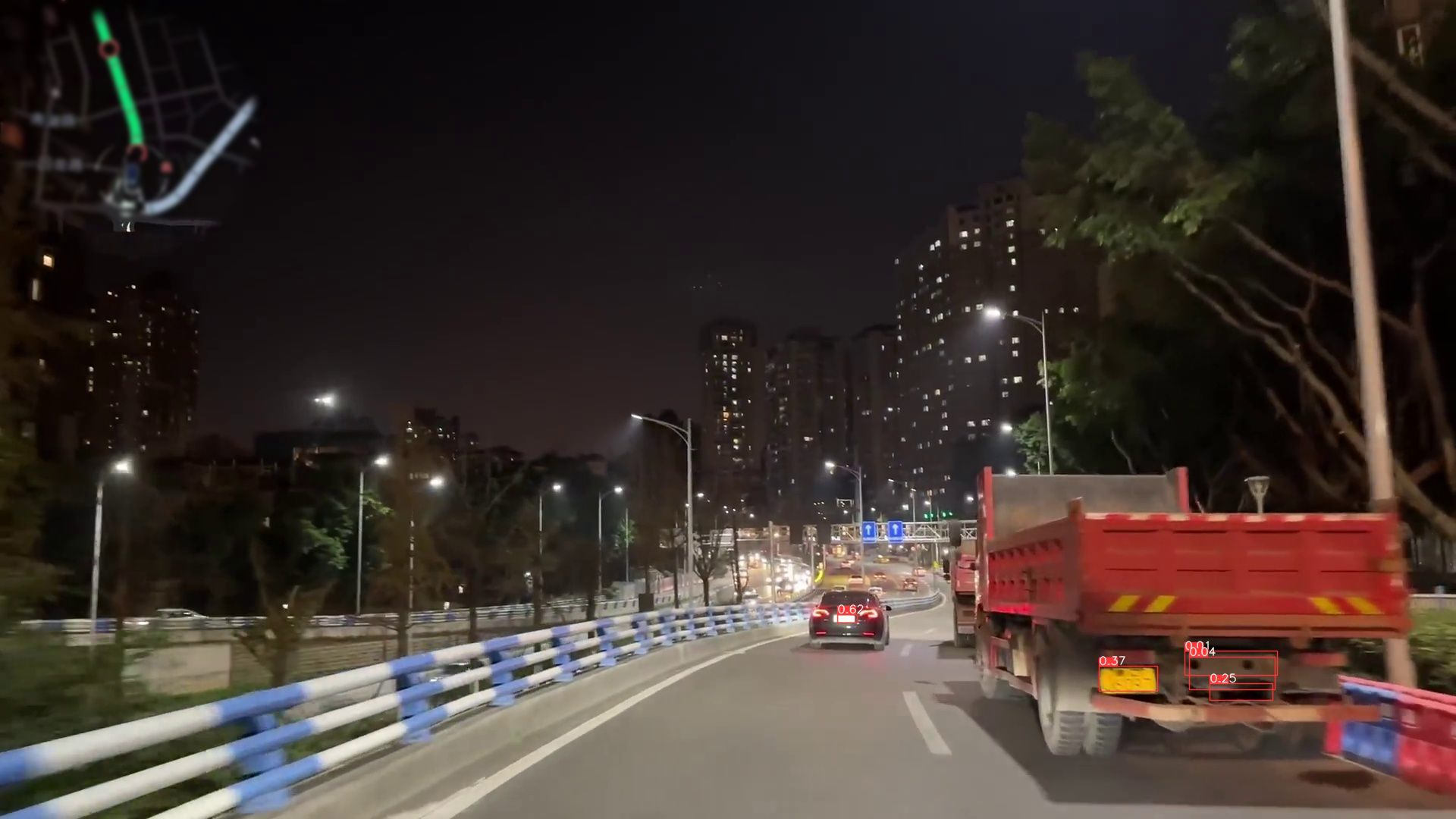}
\end{subfigure}

\vspace{1em}

\begin{subfigure}{0.47\textwidth}
    \includegraphics[width=\linewidth]{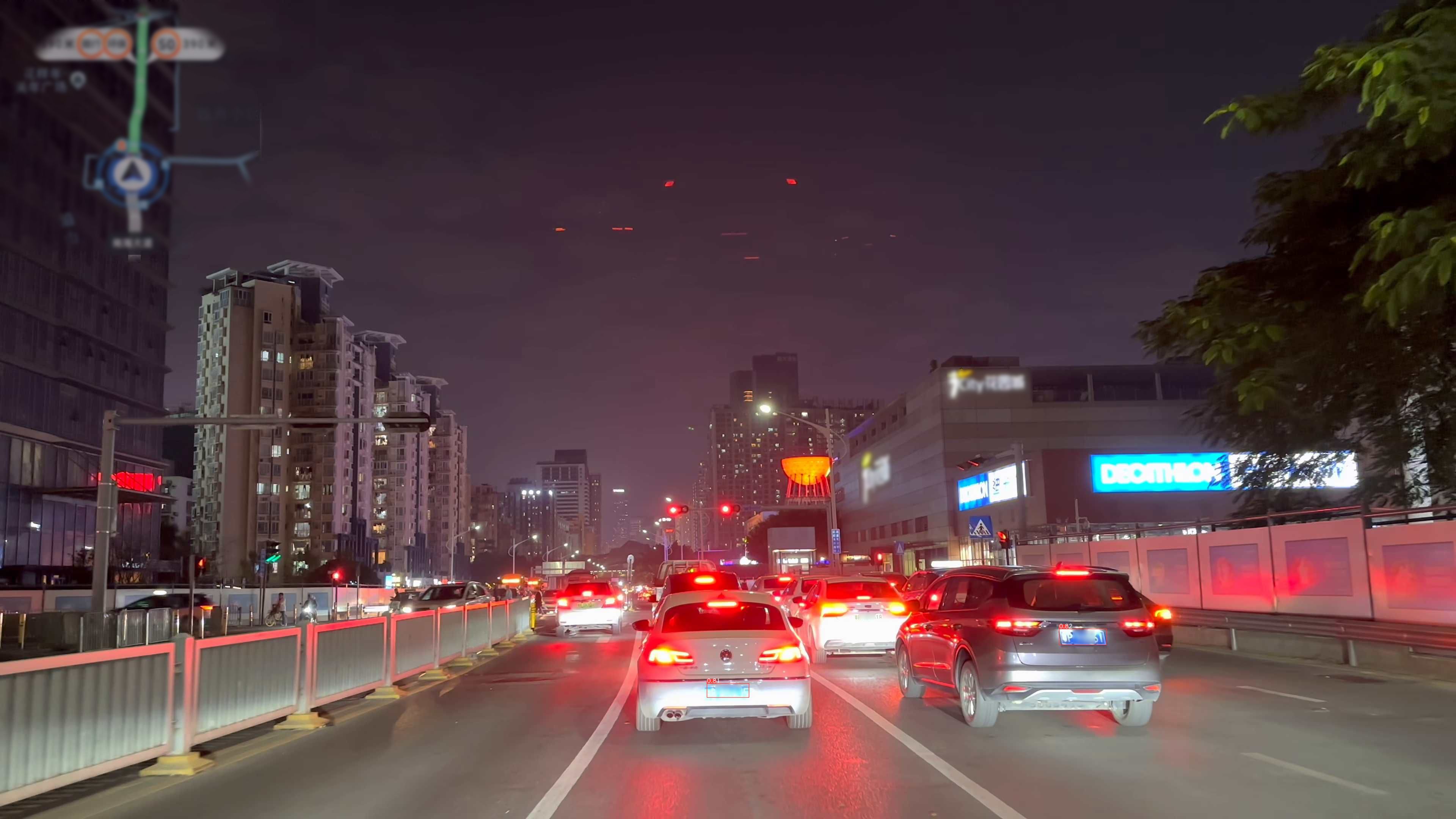}
\end{subfigure}
\hspace{0.02\textwidth}
\begin{subfigure}{0.47\textwidth}
    \includegraphics[width=\linewidth]{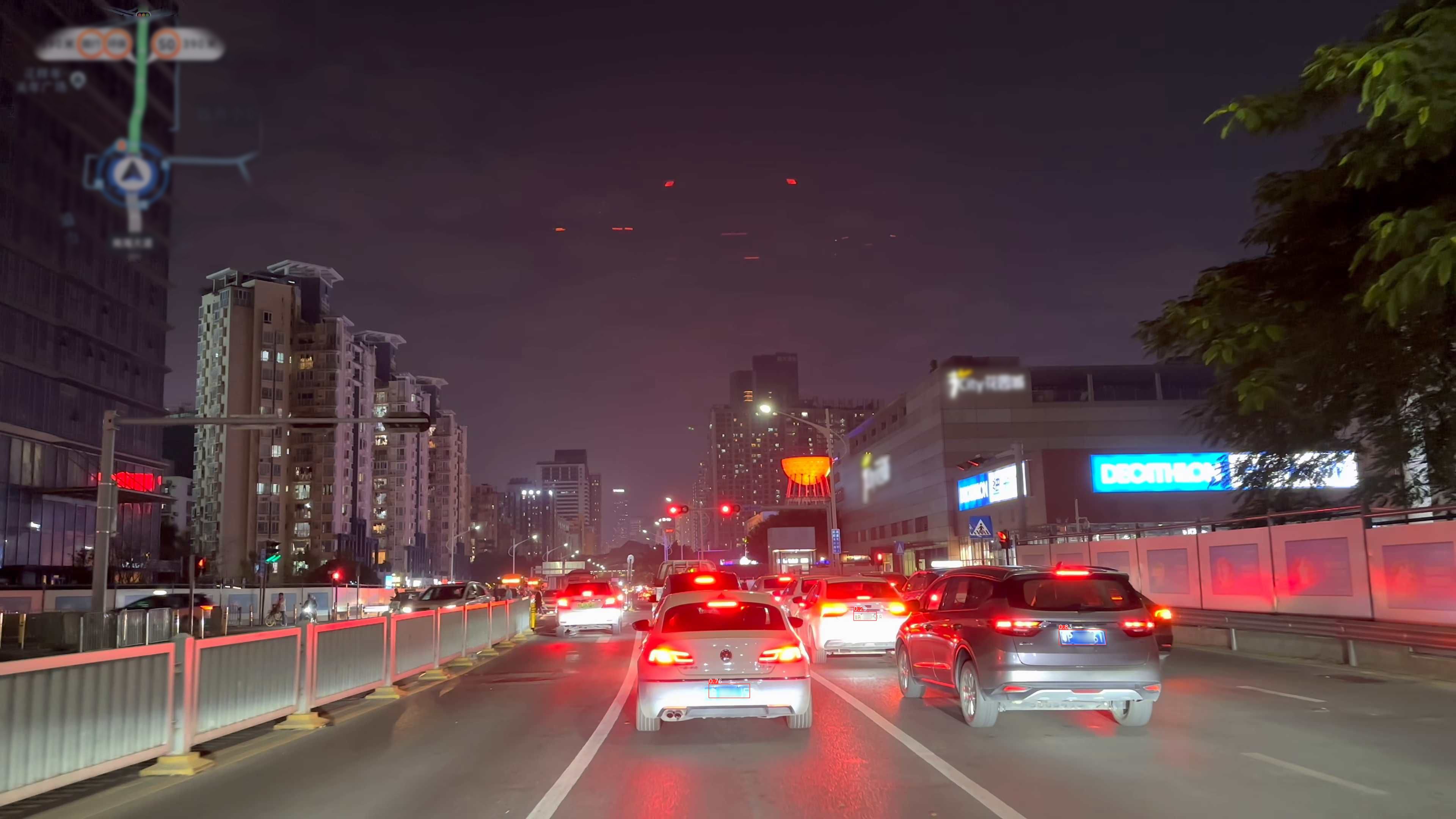}
\end{subfigure}

\caption{
Visualization of YOLOV9-T at Epoch 100 for four different images. Each row shows the CIoU loss (left) and the ICR-CIoU loss prediction (right) on SVMLP dataset.
}
\label{fig:epoch100_large}
\end{figure*}

Fig.~\ref{fig:epoch20_large} and Fig.~\ref{fig:epoch100_large} show the virtualized predicted bounding box upon different epochs, which shows that the proposed ICR loss penalty can accelerate the learning.
\vspace{-0.5em} 

\end{document}